\documentclass[lettersize,journal]{IEEEtran} 
\usepackage{arydshln}
\usepackage{amsmath,amsfonts}
\usepackage{algorithmic}
\usepackage{algorithm}
\usepackage{array}
\usepackage[caption=false,font=normalsize,labelfont=sf,textfont=sf]{subfig}
\usepackage{textcomp}
\usepackage{stfloats}
\usepackage{url}
\usepackage{verbatim}
\usepackage{graphicx}
\usepackage{cite}
\usepackage{amssymb}
\usepackage{multirow}
\usepackage{makecell}
\usepackage{colortbl}
\usepackage{threeparttable}
\usepackage[table]{xcolor}
\usepackage{tikz}
\usepackage{booktabs}
\usepackage{xcolor}
\usepackage{boldline}
\usepackage{scalefnt}
\usepackage{hyperref}
\usepackage{arydshln}
\usepackage{cancel}
\usepackage{amsmath}
\allowdisplaybreaks
\definecolor{deepgreen}{RGB}{16, 125, 82} 
\definecolor{deepblue}{RGB}{5, 94, 162} 
\definecolor{deepred}{RGB}{154, 0, 1} 

\definecolor{66b3ff}{RGB}{102, 179, 255}  
\definecolor{ffcc99}{RGB}{255, 204, 153}  

\definecolor{color1}{HTML}{FFDFBE} 
\definecolor{color2}{HTML}{FFF2E5}

\begin{document}

\title{RagGAD: Rationale-Aware Conditional Gaussian Mixture Normalizing Flow for Unsupervised Graph Anomaly Detection}

\author{Junxin~Lu, Jing Zhao, Shiliang~Sun,~\IEEEmembership{Senior Member,~IEEE}
\thanks{Junxin Lu and Jing Zhao are with the School of Computer Science
and Technology, East China Normal University, Shanghai 200062, China.
(E-mail: junxinlu.ecnu@gmail.com, jzhao2011@gmail.com)}
\thanks{Shiliang Sun is with the School of Computer Science and Technology, 
East China Normal University, Shanghai 200062, China, 
and also with the State Key Laboratory of Submarine Geoscience, School of Automation and Intelligent Sensing, Shanghai Jiao Tong University,
800 Dongchuan Road, Shanghai 200240, China. (E-mail: shiliangsun@gmail.com).}
\thanks{Corresponding author: Shiliang Sun.}}

\maketitle

\begin{abstract}
Graph anomaly detection aims to identify nodes that deviate from normal behavioral patterns within graphs. 
However, existing methods largely rely on the homophily assumption, 
which makes it difficult to distinguish spurious affinities and to capture the diverse behaviors of normal nodes,
limiting their robustness in complex real-world scenarios.
To address this problem, we propose RagGAD, an unsupervised graph anomaly detection framework based on
rationale-aware conditional Gaussian mixture normalizing flow. 
RagGAD introduces an adaptive rationale disentangler to disentangle stable rationales 
from spurious correlations within node interrelationships,
and further decomposes stable rationales into robust and fragile components.  
The learned rationales capture underlying interaction patterns that characterize normal behaviors under varying conditions, 
while anomalies emerge as deviations associated with unstable or spurious correlations.
To model the intricate distributions of normal and abnormal nodes, 
RagGAD integrates rationale-non-rationale Gaussian mixture modeling with a robust-fragile rationale mixture learning strategy. 
By mitigating spurious homophilic correlations and embracing the heterogeneity of normal patterns, 
RagGAD identifies anomalies as low-density regions within a  structure-aware distribution space.
Extensive experiments on multiple benchmark datasets demonstrate that RagGAD outperforms state-of-the-art methods.
\end{abstract}


\section{Introduction}
Graph anomaly detection (GAD) aims to identify nodes in a graph that exhibit behaviors or characteristics significantly 
deviating from the majority \cite{li2023high,qiao2024deep,li2025tered}. 
GAD is crucial in various real-world applications, such as detecting irregularities in financial \cite{li2022internet,liu2021pick} and 
social \cite{wang2022wrongdoing} networks. 
Existing GAD methods are predominantly based on supervised or semi-supervised learning paradigms to model node-level abnormalities, 
 relying on labeled data to guide model training  \cite{ding2019deep,tang2022rethinking}.
However, in real-world scenarios, data is often vast and complex, with anomalies that cannot be pre-identified or labeled.

\begin{figure}[t]
\centering
\includegraphics[width=0.49\textwidth]{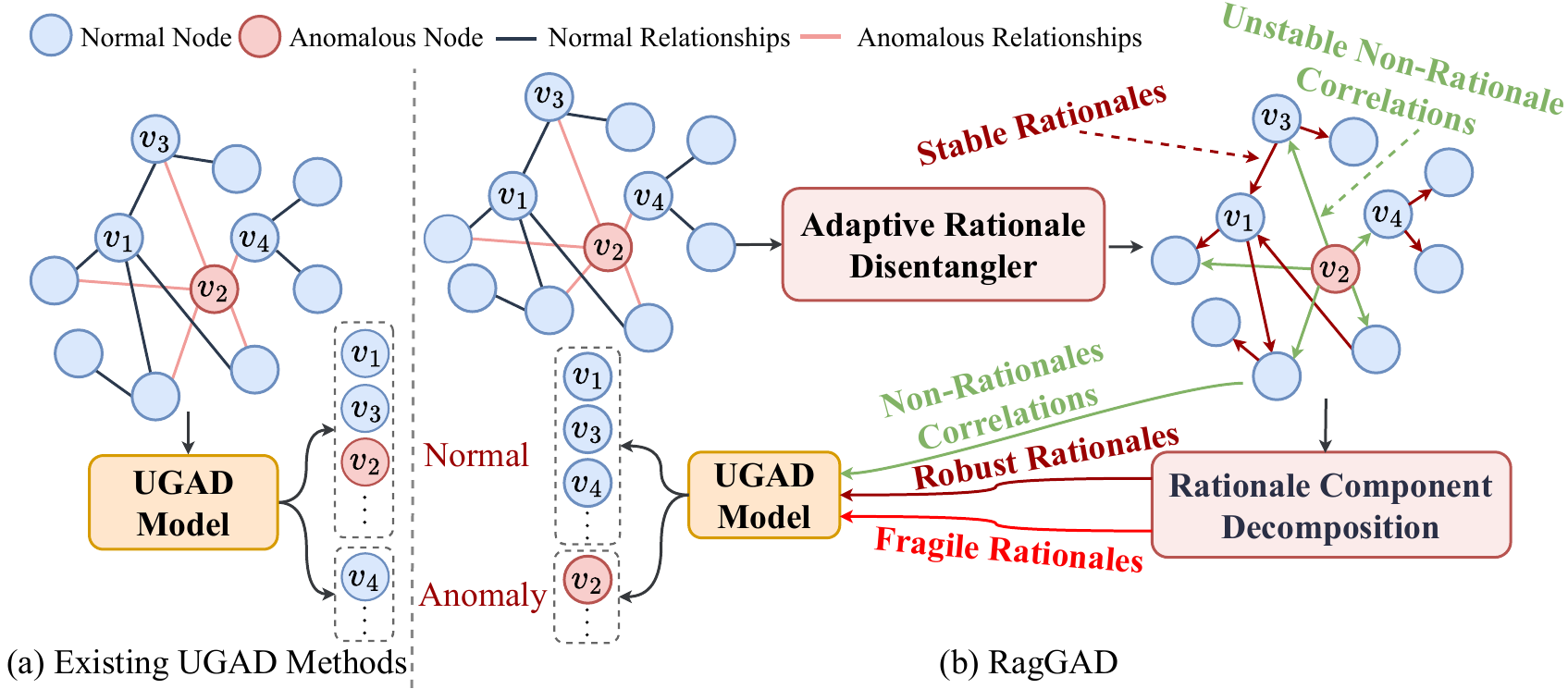} 
\vspace{-0.6cm}
\caption{In UGAD, (a) existing methods fall into the homophily trap when abnormal nodes mimic normal connection patterns (e.g., abnormal node $v_2$ 
imitating the affinity of normal node $v_1$ with its neighbors). Additionally, they incorrectly identify $v_4$ as abnormal due to overlooking the 
diversity of normal patterns. 
(b) RagGAD disentangles stable rationales,  enabling the model to capture true mutual influence while eliminate spurious connections/affinities.
Furthermore, it decomposes rationales into robust and fragile components,
 allowing for fine-grained modeling of both stable and context-sensitive normal patterns.}   
\label{fig1}
\end{figure}

In this context, unsupervised graph anomaly detection (UGAD) \cite{niu2024zero,wang2025context,qiao2024truncated} 
emerges as a more practical yet challenging paradigm, 
as it eliminates the need for labeled supervision, thereby expanding the practical applicability of GAD.
Existing UGAD works are comparatively few, which can be roughly categorized into two groups:
(i) data reconstruction-based methods, which identify anomalies by quantifying discrepancies  between reconstructed node attributes/structures  
and the original inputs \cite{he2024ada,roy2024gad,luo2022comga,dong2025smoothgnn}; 
(ii) self-supervised learning-based methods, which utilize pre-text tasks such as proxy classification, 
surrogate contrastive learning, or pre-trained models to provide additional supervisions \cite{liu2024bourne,liu2021anomaly,zheng2021generative}.

Despite achieving promising detection performance, when dealing with complex and diverse graphs, 
above-mentioned works inherently face the following limitations: 
(\emph{i}) Over-reliance on the homophily assumption, which posits that  normal nodes  have stronger affinity with each other than anomalies,
may lead to the homophily trap \cite{kipf2016semi,qiao2024truncated,he2024ada,chen2024consistency}.
When normal nodes exhibit weak affinity or abnormal nodes deliberately conceal their suspicious activity patterns,
the homophily assumption  is violated, resulting in misidentification.
As illustrated in Figure \ref{fig1}(a), the abnormal node $v_2$ disguises its abnormal behavior by mimicking the normal node $v_1$, 
sharing same neighbors and inter-nodes affinities with $v_1$. 
Consequently, $v_2$ is incorrectly identified as a normal node.
This is attributed to their neglect of deeper interaction patterns among nodes, focusing solely on statistical co-occurrence 
or superficial similarity. 
However, there exist intricate yet stable interdependencies among nodes,  referred to as  robust rationales.
These rationales capture the intricate yet stable interdependencies among nodes,
explaining the robust interaction patterns underlying specific nodes.
By modeling stable rationales, the model can uncover the genuine mutual influences between nodes,
eliminating spurious connections or affinities even when abnormal nodes obscure their behavior, as shown in Figure \ref{fig1}(b).
(\emph{ii}) In real-world graphs, nodes often exhibit diverse normal patterns, 
such as varying  preferences and personalities  within a community.  
As shown in Figure \ref{fig1}(a), nodes $v_3$ and $v_4$ are normal and connected to abnormal node $v_2$.
If the model fails to distinguish the varying directions and strengths of interactions with $v_2$, 
it may misclassify $v_4$ as abnormal. 
Therefore, the model must be sufficiently sensitive to these variations,
ensuring that diverse and fine-grained normal behaviors are not misidentified as anomalies.

\begin{figure*}[t]
\vspace{-0.4cm}
\centering
\includegraphics[width=0.9\textwidth]{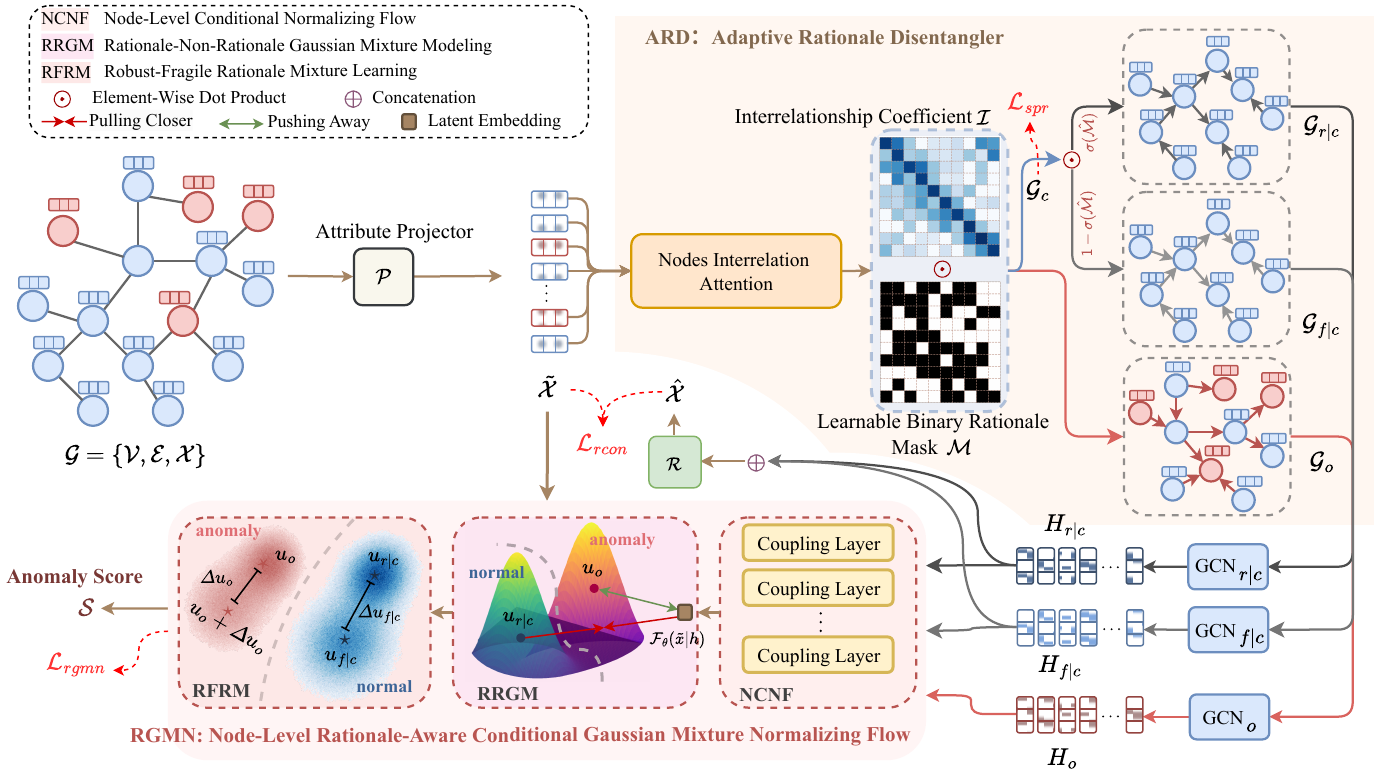} 
\vspace{-0.35cm}
\caption{Overview of RagGAD. RagGAD introduces an adaptive rationale disentangler (ARD) to disentangle rationales $\mathcal{G}_c$ and 
non-rationale correlations $\mathcal{G}_o$ from complex node interrelationships $\mathcal{I}$. 
The robust rationales $\mathcal{G}_{r|c}$ and fragile rationales $\mathcal{G}_{f|c}$ are further decomposed from $\mathcal{G}_c$
using a learnable soft mask $\hat{\mathcal{M}}$.
Based on $\mathcal{G}_{r|c}$, $\mathcal{G}_{f|c}$, and $\mathcal{G}_o$, RagGAD extracts the 
robust rationale representation $H_{r|c}$, fragile rationale representation $H_{f|c}$, and non-rationale representation $H_o$, respectively,
which are then transformed into latent conditional embeddings $z_{r|c}$, $z_{f|c}$, and $z_o$ through NCNF.
Subsequently, RRGM instantiates rationales and non-rationale correlations as distinct classes for 
Gaussian mixture modeling to fit the distribution of node attribute $\tilde{x}$ conditioned on $z_{r|c}$, $z_{f|c}$ and $z_o$.
Meanwhile, RFRM captures the latent diversification of normal distributions among nodes through 
fine-grained robust-fragile rationale Gaussian mixture learning.
}   
\label{overview}
\end{figure*}

To address these challenges, we propose \textbf{RagGAD},  a novel framework based on \underline{\textbf{Ra}}tionale-aware 
conditional \textbf{\underline{g}}aussian mixture normalizing flow for U\underline{\textbf{GAD}},  as shown in Figure \ref{overview}. 
RagGAD employs an adaptive rationale disentangler (ARD) to disentangles  stable rationale from spurious correlations in node interrelationships.
The rationales encapsulate the underlying influence mechanisms 
governing normal node behaviors, while anomalies are revealed via unstable spurious correlations.
To address the diversity of normal patterns, RagGAD employs a learnable rationale soft mask to decompose stable rationales
into robust and fragile components. 
This enables fine-grained modeling of both stable and context-sensitive normal patterns. 
Based on the disentangled stable rationales and unstable correlations, 
we propose a node-level rationale-aware conditional Gaussian mixture normalizing flow (RGMN)
model to capture complex normal-abnormal distributions, 
with anomalies identified as low-density regions within the mixture distribution.   
A key component of RGMN is rationale-non-rationale Gaussian mixture modeling (RRGM), 
which leverages conditional Gaussian mixture modeling to effectively separate 
and model the distributions of rationale and non-rationale correlations.
Additionally, to capture nuanced variations within rationale representations, 
we introduce a robust-fragile rationale mixture learning (RFRM) strategy. 
This strategy models diverse combinations of robust and fragile components, 
enabling RagGAD to learn fine-grained divergent traits across normal patterns 
and provide a comprehensive understanding of the underlying mixture distribution.
The main contributions of this paper are summarized as follows:
\begin{itemize}
  \item We disentangle stable rationales from spurious correlations within complex node interrelationships, 
  further refining rationales into robust and fragile components using an adaptive rationale disentangler.
  This enables RagGAD to capture the underlying influence mechanisms governing normal node behaviors,
  while anomalies are revealed through unstable non-rationale correlations.
  \item We propose a node-level rationale-aware conditional Gaussian mixture normalizing flow model, 
  which learns the diversity of node normal behaviors and models the intricate distributions of both normal and abnormal patterns, 
  enabling efficient anomaly detection in low-density regions.
  \item Experimental results on multiple datasets demonstrate that RagGAD significantly
   outperforms state-of-the-art baselines for unsupervised graph anomaly detection.
\end{itemize}

\section{Relative Works}
\subsection{Unsupervised Graph Anomaly Detection}
Compared to supervised \cite{tang2023gadbench,tang2022rethinking} and semi-supervised \cite{ai2025semi,huang2022auc,qiao2024lim,gao2023addressing} paradigms, 
unsupervised graph anomaly detection (UGAD) aims to identify abnormal nodes among predominantly normal nodes without 
relying on labeled data \cite{ding2019deep,luo2022comga}.

A practical and intuitive strategy is data reconstruction-based, where nodes are reconstructed based on their similarity 
or affinity with neighbors, and those with high reconstruction errors are identified as anomalies \cite{he2024ada,roy2024gad,luo2022comga,ding2019deep}.
Another line of work leverages self-supervised learning, incorporating contrastive learning 
\cite{liu2021anomaly,chen2022gccad,zheng2021generative,chen2024boosting}, proxy classification \cite{wang2021one}, and 
auxiliary objectives \cite{huang2022hop,pan2025label,pan2023prem,zhao2020error,dong2025smoothgnn}. 
Additionally, general anomaly detection methods, such as ARC \cite{liu2024arc}, UNPrompt \cite{niu2024zero} and FreeGAD \cite{zhao2025freegad},
aim to detect anomalies across domains without requiring fine-tuning and retraining, 
but suffer from low accuracy \cite{wang2023cross,liu2024arc,niu2024zero}.
FreeGAD \cite{zhao2025freegad} leverages an affinity-gated encoder and anchor-guided statistical deviations to compute anomaly scores, 
however, its reliance on heuristic anchors and shallow statistical measures limits its ability to capture complex dependencies.
ADA-GAD \cite{he2024ada} and HUGE \cite{pan2025label} alleviate the homophily trap through learning-free augmentation strategy 
and label-free heterogeneity measurement to reduce false detection rates.
CoCo \cite{wang2025context} leverages Transformers to jointly model multi-hop local and global contextual features, 
and detects anomalies by measuring discrepancies in their correlations.
GCTAM \cite{zhang2025gctam} extends TAM \cite{qiao2024truncated}, which 
identifies anomalies by maximizing normal-node similarity while truncating anomalous ones, 
by incorporating contextual information and global similarities to avoid rigid thresholds 
thereby improving the modeling of node-specific characteristics and high-order affinities.
However, they are still limited to the statistical node associations provided by the dataset, 
without further exploring more robust underlying rationale relationships among variables.

The aforementioned works typically rely on statistical co-occurrence/similarity or homophily-aware anomaly metrics 
to model intrinsic normal patterns for UGAD.
However, they struggle to eliminate spurious connections and affinities, 
making them susceptible to the homophily trap when anomalies conceal their suspicious behavior.
In contrast, RagGAD disentangles more robust dependencies, i.e., rationales, from unstable non-rationale correlations,
thereby revealing the underlying mechanisms of interactions and generative patterns among normal nodes,
 while exposing anomalies through non-rationale correlations.
 
\subsection{Normalizing Flows in Anomaly Detection}
Normalizing flows (NFs) leverage invertible transformations to map complex data distributions to simpler ones (e.g., standard normal distribution),
enabling precise probability density estimation.
NFs maximize the log-likelihood of normal samples, assuming that normal samples map to high-density regions, while anomalies fall into low-density regions 
\cite{papamakarios2021normalizing,cao2024fanfold}.
NFs have been widely explored for anomaly detection in non-graph data \cite{rudolph2021same,dai2022graph,gudovskiy2022cflow,rudolph2022fully,yao2025hierarchical}.
DifferNet \cite{rudolph2021same} uses NFs to assign meaningful likelihoods to images and develops a scoring function to detect defects.
BGAD \cite{yao2023explicit} enhances model distinguishability through a boundary-guided semi-pull contrastive learning mechanism based on NFs,
while HGAD \cite{yao2025hierarchical} introduces a hierarchical Gaussian mixture normalizing flow for unified anomaly detection.
GANF \cite{liu2019graph}  combine normalizing flow with a  graph auto-encoder to create a generative model of graph structures.
For  graph data, FANFOLD \cite{cao2024fanfold} proposes a  graph normalizing flows-driven asymmetric network for unsupervised graph-level anomaly detection.
However, how to utilize normalizing flows for the more challenging task of node-level UGAD,
while disentangling inter-node robust rationale relationships, remains unexplored in previous works. 
 
These works inspire us to leverage NFs for learning data distributions and identifying anomalies in low-density regions.
However, existing methods based on normalizing flows rely solely on normal data during training, 
which is impractical and leads to ambiguous decision boundaries between normal and abnormal data, thereby reducing distinguishability.
In contrast, RagGAD facilitates coexistence of normal and abnormal data during training, 
leveraging disentangled robust rationales and non-rationale correlations  
for discriminative rationale Gaussian mixture modeling in UGAD. 
Furthermore, existing methods overlook the diversity of normal patterns, misclassifying diverse fine-grained normal behaviors as anomalies.

\section{Preliminaries}
\subsection{Problem Statement} 
An attributed graph is represented as $G=\{\mathcal{V},\mathcal{E},\mathcal{X}\}$, where $\mathcal{V}=\{v_1, v_2, \cdots, v_N\}$ and 
$\mathcal{E}=\{\ldots, e_{ij}, \ldots\}_{i,j=1}^{N}$ are the sets of nodes and edges, respectively. 
$\mathcal{X}=\{x_i\}_{i=1}^N \in \mathbb{R}^{N \times d_x}$ is the node-level attribute matrix, 
where $x_i$ indicates the attribute vector of node $v_i$ with $d_x$ dimensions. 
The inter-node connectivity of $G$ is represented by an adjacent matrix $A=\{a_{ij}\}_{i,j=1}^N \in \mathbb{R}^{N \times N}$, 
where $a_{ij}=1$ indicates an edge between nodes $v_i$ and $v_j$, and $a_{ij}=0$ indicates no edge.

Given an attributed  graph $G=\{\mathcal{V},\mathcal{E},\mathcal{X}\}$, UGAD aims to identify abnormal nodes $\mathcal{V}_a$ 
(the minority) from normal nodes $\mathcal{V}_n$ (the majority), where  $\mathcal{V}_a \cup  \mathcal{V}_n=\mathcal{V}$,  $\mathcal{V}_a \cap  \mathcal{V}_n =  \emptyset$
and  $ \left\lvert \mathcal{V}_a \right\rvert  \ll   \left\lvert \mathcal{V}_n\right\rvert $, without access to any class labels during training.  
The objective of UGAD model is to learn an anomaly scoring function $\mathcal{S}:  \mathcal{V} \to \mathbb{R}$,
such that $\mathcal{S}(v_n) < \mathcal{S}(v_a)$ for all $\forall v_n \in \mathcal{V}_n$ and $\forall v_a \in \mathcal{V}_a$. 

\vspace{-0.2cm}
\subsection{Normalizing Flow} The normalizing flow \cite{dinh2014nice,kingma2018glow,dinh2016density} is a probabilistic density estimation model that maps 
an unknown data distribution $p(\mathcal{X})$ to a tractable latent distribution $p(Z)$  by a sequence of invertible transformations.
Specifically, a normalizing flow  model is defined as a bijective  transformation 
$\mathcal{F}: x \in \mathbb{R}^{d_x} \leftrightarrow  z \in \mathbb{R}^{d_z}$, establishing a mapping from the data space $x \in \mathcal{X}$ 
to the latent space $z \in Z$. 
According to the change of variables formula in calculus \cite{villani2021topics},  the log-likelihood of any $x \in \mathcal{X}$ can be computed as:
\begin{equation}
  \label{logpx}
  \text{log}p_{\mathcal{X}}(x) = \text{log} p_{Z}\left(\mathcal{F}_{\theta}(x)\right) + \text{log} | \text{det} J |,
\end{equation}
where  
$J=\nabla_{x}\mathcal{F}_{\theta}(x)=\frac{\delta\mathcal{F}_{\theta}(x)}{\delta x}$ is the Jacobian matrix of the transformation.
$\theta$ represents  the learnable model parameters of $\mathcal{F}$.
In most flow-based models, $p_Z(z)$ is typically assumed to follow a standard multivariate normal distribution $\mathcal{N}(0, \mathbb{I})$ for simplicity.  
The normalizing flow $\mathcal{F}_{\theta}$ can be optimized by maximizing the log-likelihood over the training distribution $p(\mathcal{X})$.
Consequently, the loss function is defined as follows:
\begin{equation}
  \mathcal{L}_{nf} = \mathbb{E}_{x \sim p(\mathcal{X})}[-\text{log}p_{\mathcal{X}}(x)].
\label{eq1}
\end{equation}
Conditional normalizing flow \cite{ardizzone2019guided,zhou2024label,abdelhamed2019noise} extends normalizing flow by introducing an external condition 
$\mathcal{C}$, allowing the model to estimate the conditional distribution 
$p_{\mathcal{X}}(x|\mathcal{C})=p_{Z}\left(\mathcal{F}_\theta(x;\mathcal{C})\right) \left\lvert\text{det}\nabla_x \mathcal{F}_\theta(x;\mathcal{C}) \right\rvert $, 
thereby enabling conditional density estimation. 
The log-density of $x$ conditioned on $\mathcal{C}$ is defined as follows:
\begin{equation}
   \text{log} p_{\mathcal{X}}(x|\mathcal{C})=\text{log}p_Z\left(\mathcal{F}_\theta(x;\mathcal{C})\right) + \text{log}\left\lvert \text{det} \nabla_x \mathcal{F}_\theta(x;\mathcal{C})  \right\rvert. 
\label{eq3}
\end{equation}

\section{Methodology}
We propose RagGAD, a node-level unsupervised graph anomaly detection framework 
based on rationale-aware conditional Gaussian mixture normalizing flow. 
\subsection{Adaptive Rationale Disentanglement} \label{ARD}
In UGAD, only the node attribute matrix $\mathcal{X}$ and the adjacent matrix $A$ are accessible during training, 
while the stable rationales $\mathcal{G}_{c}$ and unstable non-rationale correlations $\mathcal{G}_{o}$ are unknown.
To tackle this, we propose an adaptive rationale disentanglement module, which effectively disentangles the
$\mathcal{G}_{c}$ and $\mathcal{G}_{o}$ from node interdependencies $\mathcal{I}$.   

RagGAD first employs an attribute projector $\mathcal{P}(\cdot)$ to reduce the attribute dimension of nodes \cite{zhao2024all,liu2024arc}.
Specifically, given the attribute matrix $\mathcal{X} \in \mathbb{R}^{N \times d_x}$, the attribute projection is defined as follows:
\begin{equation}
  \tilde{\mathcal{X}} \in \mathbb{R}^{N \times d_{\tilde{x}}} = \mathcal{P}(\mathcal{X}) =  \mathcal{X}W_{\mathcal{P}},
  \label{projection}
\end{equation}
where $\tilde{\mathcal{X}}$ is the projected attribute matrix.
$W_{\mathcal{P}} \in \mathbb{R}^{N \times d_{\tilde{x}}}$ is a dataset-specify linear projection weight matrix, and $d_{\tilde{x}}$ is a 
pre-defined dimension of the projected attribute for nodes.   
The attribute projector $\mathcal{P}(\cdot)$ can employ commonly used dimensionality reduction methods, 
such as singular value decomposition \cite{stewart1993early} or principal component analysis \cite{abdi2010principal}.

\hspace{1em}{\emph{a) Adaptive Rationale Disentangler.} 
We propose an adaptive rationale disentangler (ARD) to capture the stable rationales 
$\mathcal{G}_c$ among nodes, while simultaneously 
disentangling robust rationales $\mathcal{G}_{r|c}$ and fragile rationale relationships $\mathcal{G}_{f|c}$. 
Specifically, ARD integrates a node interrelation attention module to learn the structure interrelationships between nodes,  
forming the node interrelationships  $\mathcal{I}$.
This involves calculating the cross attention coefficient $\varepsilon_{ij}$, 
which reflects the interrelationship of node $v_j$ to $v_i$, defined as:
\begin{equation}
  \varepsilon_{ij} = \frac{\mathbb{I}\left[a_{ij}=1\right]\cdot\text{exp}\left(Q_iK_j^T/\sqrt{d_{\tilde{\mathcal{X}}}}\right)}
  {\sum_k\mathbb{I}\left[a_{ik}=1\right]\cdot\text{exp}\left(Q_iK_k^T/\sqrt{d_{\tilde{\mathcal{X}}}}\right)},
\end{equation} 
where $Q=\tilde{x}_i W_Q \in \mathbb{R}^{d_{\tilde{x}}}$ is the query embedding of $v_i$, 
and $K=\tilde{x}_j W_K \in \mathbb{R}^{d_{\tilde{x}}}$ is the key embedding of $v_j$. 
$W_Q \in \mathbb{R}^{d_{\tilde{x}} \times d_{\tilde{x}}}$ and $W_K\in \mathbb{R}^{d_{\tilde{x}} \times d_{\tilde{x}}}$ 
are the query and key weights, respectively. 
According to the adjacent matrix $A$, node $v_j$ is constrained to be a neighbor of node $v_i$, 
i.e., $v_j \in \mathcal{N}(v_i)=\{v_k| a_{ik}=1\}$. 

RagGAD treats the disentanglement of $\mathcal{G}_c$ and  $\mathcal{G}_o$ as an edge selection problem. 
Some edges reflect stable underlying dependencies between nodes, i.e., rationales, while others represent unstable relationships determined 
by proximity or statistical correlations.
Therefore, we use a binary mask as disentangler, denoted as $\mathcal{M}=\{m_{ij}\}_{i,j=1}^N \in \{0,1\}^{N \times N}$,
as shown in Figure \ref{overview}. 
This allows us to disentangle $\mathcal{G}_c \in \mathbb{R}^{N \times N}$ and $\mathcal{G}_o \in \mathbb{R}^{N \times N}$
from $\mathcal{I}=\{\varepsilon_{ij}\}_{i,j=1}^N$ as follows:
\begin{equation}
\begin{cases} 
  \mathcal{G}_c = \mathcal{I} \odot  \mathcal{M}, \\
  \mathcal{G}_o = \mathcal{I} \odot  \left(1 - \mathcal{M}\right),    
\end{cases} \; \text{s.t.}  \; \mathcal{M} = \mathbb{I}_{\tilde{m}_{ij} > \tau } \sigma(\tilde{\mathcal{M}}),
\label{causal_d}
\end{equation}
where $\tilde{\mathcal{M}}=\{\tilde{m}_{ij}\}_{i,j=1}^N \in \mathbb{R}^{N \times N}$ is the trainable mask of $\mathcal{M}$. 
$\sigma(\cdot)$ is the sigmoid function,  $\tau$ is the hyperparameter for selecting rationale edges, 
and $\odot$ represents element-wise multiplication. 

In UGAD, relying solely on the disentanglement of rationales and non-rationale correlations remains insufficient
for fully capturing the complex and diverse patterns underlying normal nodes.
This limitation may cause certain nodes, which exhibit diverse behaviors but are inherently normal, 
to be mistakenly identified as anomalies.
To address this, we further decompose $\mathcal{G}_c$ into stable rationales $\mathcal{G}_{r|c}$ and 
fragile rationales $\mathcal{G}_{f|c}$. 
$\mathcal{G}_{r|c}$ captures persistent, robust relationships,   
while $\mathcal{G}_{f|c}$ reflects context-sensitive relationships that are more susceptible to data perturbations or neighborhood fluctuations.
To achieve this decomposition, RagGAD introduce a learnable soft mask $\hat{\mathcal{M}} \in \mathbb{R}^{N \times N}$, 
which partitions $\mathcal{G}_c$ into $\mathcal{G}_{r|c}$ and $\mathcal{G}_{f|c}$ as follows:
\begin{equation}
  \mathcal{G}_{r|c} = \mathcal{G}_c \odot \sigma(\hat{\mathcal{M}}), 
  \quad  \mathcal{G}_{f|c} = \mathcal{G}_c \odot (1 - \sigma(\hat{\mathcal{M}})),
\end{equation}
where $\hat{\mathcal{M}}=\{\hat{m}_{ij}\}_{i,j=1}^N$, with $\hat{m}_{ij} \in [0,1]$ representing the learnable 
weight for the stability of the rationales from node $v_j$ to node $v_i$.
This decomposition enables a fine-grained re-weighting partitioning of $\mathcal{G}_c$, 
where each rationale edge is assigned a stability score via $\sigma(\hat{\mathcal{M}})$.  
By combining diverse robust and fragile rationales, RagGAD effectively delineates the boundaries between 
stable and fragile rationale relationships, 
offering a nuanced understanding of the underlying anomalies within the graph while comprehensively 
capturing the behavioral diversity of normal nodes.

Upon extracted $\mathcal{G}_{r|c}$ and $\mathcal{G}_{f|c}$, 
we introduce two rationale augmented graph convolution layers,
$\text{GCN}_{r|c}(\cdot)$ and $\text{GCN}_{f|c}(\cdot)$, followed by a reconstructor  $\mathcal{R}(\cdot)$, 
to generate the node attribute reconstruction.  The reconstruction process is defined as follows:
\begin{equation} 
\footnotesize
\begin{split}
  \hspace{-0.5cm} & \hat{\mathcal{X}}  = \mathcal{R}\left(\left[H_{r|c}\|H_{f|c}\right];\theta_\mathcal{R}\right), \\
  \hspace{-0.5cm}   H_{r|c}\hspace{-0.1cm}= & \text{GCN}_{r|c}(\mathcal{G}_{r|c},\tilde{\mathcal{X}}) \hspace{-0.05cm}=W^2_{r|c} \hspace{-0.05cm}\left( \hspace{-0.05cm}\text{ReLU} \hspace{-0.05cm}\left(\mathcal{G}_{r|c}\tilde{\mathcal{X}}W^1_{r|c}\right)\hspace{-0.05cm}\right)  \hspace{-0.1cm}+ \hspace{-0.05cm}b^2_{r|c}, \\
  \hspace{-0.5cm}   H_{f|c}\hspace{-0.1cm}= & \text{GCN}_{f|c}(\mathcal{G}_{f|c},\tilde{\mathcal{X}}) \hspace{-0.05cm}=W^2_{f|c} \hspace{-0.05cm}\left( \hspace{-0.05cm}\text{ReLU} \hspace{-0.05cm}\left(\mathcal{G}_{f|c}\tilde{\mathcal{X}}W^1_{f|c}\right)\hspace{-0.05cm}\right)  \hspace{-0.1cm}+ \hspace{-0.05cm}b^2_{f|c},
\end{split}
\end{equation} 
where $W^1_{*|c} \in \mathbb{R}^{d_{\tilde{x}} \times d_{\tilde{x}}}$,  
$W^2_{*|c} \in \mathbb{R}^{d_{\tilde{x}} \times d_h}$, and $b^2_{*|c} \in \mathbb{R}^{d_h}$ 
are the learnable parameters of $\text{GCN}_{r|c}(\cdot)$ and $\text{GCN}_{f|c}(\cdot)$, respectively.
Inspired by the success of GCNs in node representation learning  through the aggregation of neighborhood effects
\cite{suncaudits,wu2025mathsf}, $\text{GCN}_{r|c}(\cdot)$ and $\text{GCN}_{f|c}(\cdot)$ aggregate  the attributes of parent nodes to
encode dependency relationships, yielding node-level rationale representations $H_{r|c} \in \mathbb{R}^{N \times d_h}$ 
and $H_{f|c} \in \mathbb{R}^{N \times d_h}$.
The $\left[\cdot\|\cdot\right]$ denotes the concatenation of representations.
$\mathcal{R}(\cdot)$, parameterized by $\theta_{\mathcal{R}}$,  is a MLP that 
transforms the fused rationale representation $H=\left[H_{r|c}\|H_{f|c}\right]$ into the reconstructed node attributes $\hat{\mathcal{X}}$.
Similarity, another graph convolution layer, $\text{GCN}_o(\cdot)$, is employed to extract the non-rationale 
representation $H_o \in \mathbb{R}^{N \times d_h}$, i.e., $H_o=\text{GCN}_o(\mathcal{G}_o,\tilde{\mathcal{X}})$. 

\vspace{-0.3cm}
\subsection{Rationale-Aware Node-Level Gaussian Mixture Normalizing Flow Model}
Inspired by the success of anomaly detection strategies based on normalizing flow,
we leverage normalizing flow to learn a robust anomaly detection criterion for UGAD.
Normalizing flow maps data into learned distributions, with the assumption that abnormal instances tend to reside in low-density regions, 
while normal instances are concentrated in high-density regions \cite{zhou2023detecting,wang2020deep,cao2024fanfold}. 
However, applying normalizing flow to UGAD presents several challenges: 
\emph{(i)}  how to accurately estimating the density of a mixture distribution containing both normal and abnormal data,
thereby distinguishing high-density (normal) and low-density (abnormal) regions; 
\emph{(ii)} how to capture subtle variations in rationale representations to enable the model to learn the diverse patterns inherent in normal nodes, 
providing a more nuanced understanding of the underlying mixture distribution.

To address these challenges, we propose a node-level rationale-aware conditional Gaussian mixture normalizing flow model (RGMN), 
as illustrated in Figure \ref{overview}. 
RGMN comprises three core modules: 
(i) a \emph{Node-Level Conditional Normalizing Flow (NCNF)} block,  
which conditionally transforms both rationale and non-rationale node representations into latent conditional embeddings,
thereby facilitating conditional density estimation; 
(ii) a \emph{Rationale-Non-Rationale Gaussian Mixture Modeling (RRGM)} module,  
which separates and models rationale and non-rationale distributions using conditional Gaussian mixture modeling, 
enabling a more precise distinction between normal and abnormal nodes;
and (iii) a \emph{Robust-Fragile Rationale Mixture Learning (RFRM)} strategy, which learns the  
 Gaussian mixture distribution of rationale representations, 
 capturing fine-grained divergent patterns among normal nodes and characterizing their diverse traits.

\hspace{1em}{\emph{a) Node-Level Conditional Normalizing Flows.}
\label{ncnf} 
The normalizing flow $\mathcal{F}$ in RagGAD is built upon the Real-NVP \cite{dinh2016density},
comprising $\eta$ affine coupling layers with an identical structure, as shown in Figure \ref{overview}.
 Specifically,  given the robust rationale representation $h_{r|c}^i \in \mathbb{R}^{d_h}$, 
the affine transformation that maps the projected attribute $\tilde{x}_i \in \mathbb{R}^{d_{\tilde{x}}}$ of node $v_i$, 
conditioned on $h_{r|c}^i$ (i.e., condition $\mathcal{C} = h_{r|c}^i$), to the latent conditional embedding $z_{r|c}^i \in \mathbb{R}^{d_z}$ is defined as:
\begin{align}
    \hspace{-0.3cm} &  z_{r|c}^i=\mathcal{F}_\theta(\tilde{x}_i ; h_{r|c}^i)  \Rightarrow   z_{r|c}^i = \left[z_{r|c}^{i,1}\|z_{r|c}^{i,2}\right], \\
    \hspace{-0.3cm} & \tilde{x}_i^1, \tilde{x}_i^2 = \Theta(\tilde{x}_i),  z_{r|c}^{i,1}=\tilde{x}_i^1,  \\
    \hspace{-0.3cm}   z_{r|c}^{i,2}= \tilde{x}_i^2&\odot \text{exp}\hspace{-0.1cm}\left(\hspace{-0.1cm}s\hspace{-0.1cm}\left(\hspace{-0.05cm}\left[\tilde{x}_i^1 \|h_{r|c}^i W_{h} \hspace{-0.05cm}\right]\right)\right) 
    + t\left(\hspace{-0.05cm}\left[\tilde{x}_i^1 \| h_{r|c}^i W_{h}\hspace{-0.05cm}\right] \right),  
\end{align} 
where $\Theta$ is a attribute partitioning function \cite{ardizzone2019guided} that splits $\tilde{x}_i$ into 
$\tilde{x}_i^1\in \mathbb{R}^{d_{\tilde{x}/2}}$ and $\tilde{x}_i^2 \in \mathbb{R}^{d_{\tilde{x}/2}}$, respectively. 
$s(\cdot)$ and $t(\cdot)$ are transformation coefficients predicted by a learnable neural network,
and is implemented using a MLPs \cite{dinh2016density,yao2025hierarchical}. 
$W_h \in \mathbb{R}^{d_{h} \times d_{\tilde{x}/2}}$ is a learnable parameters, and $\odot$ represents the element-wise product. 
Based on Eq.(\ref{eq3}), the robust rationale conditional density of $\tilde{x}_i$ can be defined as:
\begin{equation}
    \hspace{-0.4cm}  \text{log} p_{\mathcal{X}}(\tilde{x}_i|h_{r|c}^i) \hspace{-0.1cm} = \hspace{-0.1cm} 
    \text{log}p_Z \hspace{-0.1cm} \left(\hspace{-0.1cm}\mathcal{F}_\theta(\tilde{x}_i;h_{r|c}^i)\hspace{-0.1cm}\right) \hspace{-0.1cm}  + \hspace{-0.1cm}
    \text{log}\hspace{-0.1cm}\left\lvert \hspace{-0.05cm}\text{det} \nabla_x \mathcal{F}_\theta(\tilde{x}_i;h_{r|c}^i)\hspace{-0.05cm}\right\rvert. 
\label{logpx}
\end{equation}
Similarly,  conditioned on the fragile rationale representation $h^i_{f|c}$ and the non-rationale representation $h_o^i$,
we can  obtain the  latent conditional embeddings  $z_{f|c}^i$ and $z_{o}^i$ for node $v_i$, 
along with the corresponding fragile rationale conditional density  $\text{log} p_{\mathcal{X}}(\tilde{x}_i|h_{f|c}^i)$ 
and non-rationale conditional density $\text{log} p_{\mathcal{X}}(\tilde{x}_i|h_{o}^i)$.
For simplicity, we omit the superscript $i$ in the subsequent sections without loss of generality.

\hspace{1em}{\emph{b)  Rationale-Non-Rationale Gaussian Mixture Modeling.}
Based on NCNF (\ref{ncnf}), we can derive the conditional densities $\text{log} p_{\mathcal{X}}(\tilde{x}_i|h_{r|c}^i)$,
$\text{log} p_{\mathcal{X}}(\tilde{x}_i|h_{f|c}^i)$, and $\text{log} p_{\mathcal{X}}(\tilde{x}_i|h_{o}^i)$ 
for $\tilde{x}$ of node $v_i$ under the representations $h_{r|c}^i$, $h_{f|c}^i$, and $h_{o}^i$, respectively. 
However, existing normalizing flow models typically adopt a standard multivariate normal distribution $\mathcal{N}(0, \mathbb{I})$ as the 
prior and assume that all latent conditional embeddings $z$ follow an identical distribution. 
This assumption is inherently restrictive in UGAD, as it fails to capture the distinct characteristics of 
normal and abnormal nodes in the latent space,
thereby weakening the ability to distinguish between normal and abnormal patterns.

Therefore, to better fit the normal distribution in the presence of anomalies,  
we introduce rationale-non-rationale Gaussian mixture modeling (RRGM), 
which models rationale and non-rationale representations as distinct latent classes.
Specifically, RRGM formulates a conditional Gaussian mixture model based on the dependencies among representation classes, 
where rationale representations correspond to the normal class (i.e., label $Y=0$)  and 
non-rationale representations as the abnormal class (i.e., label $Y=1$). 
Consequently, RagGAD uses class-dependent mean $u_y$ and covariance $\varSigma_y$ for each class 
as the prior distribution for $z$, defined as:
\begin{equation}
\label{eq9}
\begin{split}
\hspace{-0.4cm}  & p_Z(z) = \hspace{-0.3cm}\sum_{y \in \{0,1\}}\hspace{-0.2cm} p(y)\mathcal{N}\left(z; u_y, \varSigma_y\right),  \\
\hspace{-0.4cm}  & p_{Z|Y}(z|y) = \mathcal{N}(z; u_y, \varSigma_y),  
\end{split}
\end{equation}
where $u_y$ and $\varSigma_y$ are the mean and covariance matrix of class $y$, respectively. 
$p(y)=\text{softmax}\left(W_y^T z + b_y\right)$ represents the class prior distribution of $z$,
where $W_y \in \mathbb{R}^{d_z}$ and $b_y \in \mathbb{R}$ are the learnable parameters.
RRGM setting the covariance matrix $\varSigma_y$ is the identity matrix $\mathbb{I}$ for better convergence.
Then, we can calculate the log-likelihood for latent conditional embedding $z$ as follows:
\begin{align}
    & \text{log}p_Z(z) = \text{log}\left[\sum_{y \in\{0,1\}} p(y) \mathcal{N}(z;u_y,\varSigma_y)\right]  \\
    & = \text{log}\left[\sum_{y \in\{0,1\}} p(y) \mathcal{N}(z;u_y, \mathbb{I} )\right]  \\
    & \Downarrow  p_Z(z)=(2\pi)^{-\frac{d_{z}}{2}}e^{-\frac{1}{2}}  z^Tz,     \\ 
    & =  -\frac{d_{z}}{2} \text{log}(2\pi) + \text{log}\left(\sum_{y  \in\{0,1\}} e^{\epsilon_y}\cdot e^{-\frac{\left\lVert z- u_y \right\rVert^2_2 }{2}}\right) \\
    & =  -\frac{d_{z}}{2} \text{log}(2\pi) +  \text{log}\left(\sum_{y  \in\{0,1\}} e^{-\frac{\left\lVert z- u_y \right\rVert^2_2}{2}+ \epsilon_y}     \right)  \\
    & = -\frac{d_{z}}{2} \text{log}(2\pi) + {\underset{y \in\{0,1\}} {\text{logsumexp}}}   \left(  -\frac{\left\lVert z- u_y \right\rVert^2_2}{2} + \epsilon_y \right),  
\end{align}
where $\epsilon_y=\text{log}p(y)$ denotes the logarithmic class weight, scaling the class prior distribution. 
Further, we bring $\text{log}p_Z(z)$ into Eq. (\ref{logpx}), and reformulate the conditional log-density 
$\text{log} p_{\mathcal{X}} (\tilde{x}|h)$ of $\tilde{x}$ as follows:
\begin{equation}
  \begin{split}
    \hspace{-0.3cm}   \text{log} p_{\mathcal{X}} (\tilde{x}|h) & =  {\underset{y} {\text{logsumexp}}}   
     \left(  -\frac{\left\lVert \mathcal{F}_\theta(\tilde{x};h)  - u_y \right\rVert^2_2}{2} + \epsilon_y \right) \\
     \hspace{-0.3cm}  &  -\frac{d_{z}}{2} \text{log}(2\pi)  + \text{log}|\text{det}J|, \\ 
   \end{split}
   \label{eq19}
 \end{equation}
where $J=\nabla_x \mathcal{F}_\theta(\tilde{x};h)$. 
The loss function of RRGM is defined as follows:
\begin{equation}
\small
\begin{split}
\hspace{-0.4cm}      \mathcal{L}_{rrgm}   & = \mathbb{E}_{\tilde{x} \sim p(\mathcal{X})}[-\text{log}p_{\mathcal{X}}(\tilde{x}|h)] \\
\hspace{-0.4cm}  & = \mathbb{E}_{\tilde{x} \sim p(\mathcal{X})}
[-\underset{y}{\text{logsumexp}}\hspace{-0.1cm} \left(\hspace{-0.1cm} -\frac{\left\lVert \mathcal{F}_\theta(\tilde{x};h)  - u_y  \right\rVert_2^2}{2} +   \epsilon_y \hspace{-0.1cm} \right) \\
\hspace{-0.4cm}   & - \text{log}\left\lvert\text{det}J\right\rvert  + \frac{d_z}{2}\text{log}(2\pi)].   
\end{split}
\label{rrgm}
\end{equation}

\hspace{1em}{\emph{c) Robust-Fragile Rationale Mixture Learning.}
In real-world large-scale graphs, the diversification of normal patterns among nodes is a critical problem that must be considered for UGAD. 
If a normalizing flow model can learn the diverse fine-grained normal distributions of nodes,  
it can help suppress false detections for divergent patterns within normals.  
Consequently, we propose RFRM strategy, which captures the diversity among normals nodes through fine-grained rationale Gaussian mixture learning.
Specifically, we first extends the Gaussian prior $p_{Z|Y}(z|y) = \mathcal{N}(z; u_y, \varSigma_y)$ to mixture Gaussian prior 
$p_{Z|Y}(z|y) = \sum_{k=1}^K p_k(y)\mathcal{N}(z; u_y^k, \varSigma_y^k)$, 
where $K$ is the number of Gaussian components \cite{izmailov2020semi,yao2025hierarchical}.
If, we further assume $\varSigma_y^k=\mathbb{I}$, then the likelihood of latent conditional embedding $z$ can be calculate as follows:
\begin{equation}
  p_Z(z)=\sum_y p(y) \left( \sum_{k=1}^K  p_k(y)  \mathcal{N}(z; u_y^k, \varSigma_y^k)\right).
\end{equation}
When $p_Z(z)$ is obtained, the log-likelihood $\text{log}p_Z(z)$ of $z$ can be defined as follows:
\begin{equation}
\footnotesize
\hspace{-0.4cm}  \text{log}p_Z(z) = \text{log} \hspace{-0.1cm}   \left( \hspace{-0.1cm}   \sum_y p(y)   {\underset{k} { \text{sumexp}}}   \hspace{-0.1cm}  
                  \left[ -\frac{\left\lVert z - u^k_y \right\rVert^2_2 }{2}
                  + \epsilon^k_y \hspace{-0.1cm}   -  \hspace{-0.1cm}   \frac{d_{z}}{2} \text{log}(2\pi) \right]   \hspace{-0.1cm}    \right),  
\end{equation}
where the latent conditional embedding $z=\mathcal{F}_\theta(\tilde{x};h)$. $\epsilon^k_y=\text{log}p_k(y)=\text{logsoftmax}_k(\psi_y)$ denotes the logarithmic component center weights, 
and $\psi_y$ is a learnable vector specify for class $y$, with $\psi_y \in \mathbb{R}^K$,  to adaptively learn the component weights. 
Further, we can reformulate Eq. (\ref{eq19}) to get the conditional log-density $\text{log}p_{\mathcal{X}}(\tilde{x}|h)$ 
for any $\tilde{x}$ on condition $h$ when upon mixture Gaussian prior:
\begin{equation}
\footnotesize
\begin{split}
\hspace{-0.45cm}  \text{log} p_{\mathcal{X}}(\tilde{x}|h) &  \hspace{-0.1cm} =  \hspace{-0.1cm} \text{log}  \hspace{-0.1cm} \left( \hspace{-0.1cm} \sum_{y} p(y) {\underset{k} { \text{sumexp}}}  
\hspace{-0.1cm} \left[ -\frac{\left\lVert z - u^k_y \right\rVert^2_2 }{2}
+ \epsilon^k_y - \frac{d_{z}}{2} \text{log}(2\pi)  \right]  \hspace{-0.1cm} \right)   \\ 
\hspace{-0.4cm}   & + \text{log}|\text{det}J|. \\ 
\end{split}
\label{eq17}
\end{equation}

Furthermore, in RFRM, for each class $y$, we learn a central component $u_y^c$ and a set offsets $\{\varDelta u_y^k\}_{k=1}^K$, 
used to calculate the centers $u_y^k = \{u_y^c + \varDelta u_y^k\}_{k=1}^K$  for other components. 
We can directly optimize the central component $u_y^c$  using Eq. (\ref{rrgm}) for each class $y$.  
However, when optimizing $u_y^k$ for other components, we detach the gradient of  $u_y^c$ and 
optimize the  offset $\varDelta u_y^k$ by the following loss function:    
\begin{equation}
\small
  \begin{split}
  \hspace{-0.4cm}   \mathcal{L}_{rfrm}  & = \mathbb{E}_{(\tilde{x}, y) \sim p(\mathcal{X}, Y)}[-\text{log}p_{\mathcal{X}}(\tilde{x}|h)]  \\
  \hspace{-0.4cm}  & = \mathbb{E}_{(\tilde{x}, y) \sim p(\mathcal{X}, Y)} \left[ - \text{log}\left\lvert\text{det}J\right\rvert \right.  \\  
  \hspace{-0.45cm} &  \left. - \underset{k}{\text{logsumexp}} \hspace{-0.1cm}\left( \hspace{-0.1cm}
  -\frac{\left\lVert z - \left(\cancel{\mathbf{G}}\hspace{-0.05cm}\left[u_y^c\right]  + \varDelta u^k_y \right) \right\rVert_2^2}{2} 
  + \epsilon_y^k \right) \right],
  \end{split}
\label{rfrm}
\end{equation}
where  $\cancel{\mathbf{G}}\left[\cdot\right]$ is stop gradient back-propagation.  
$\frac{d_{z}}{2} \text{log}(2\pi)$ is a constant term that does not affect the gradient during model optimization, so it is omitted.

For normal class (i.e., $Y=0$), to model the diverse combinations of robust rationales $\mathcal{G}_{r|c}$ and 
fragile rationales $\mathcal{G}_{f|c}$, 
we assign rationale latent conditional embedding $z_{r|c}$ as central component and fragile rationale latent conditional embedding $z_{f|c}$
as a complementary component.
Specifically, we initialize the central component  as  $u_{r|c}$ for $z_{r|c}$, and 
the learnable offset $\varDelta u_{f|c}$ for $z_{f|c}$ is defined as follows:
\begin{equation}
\small
  \varDelta u_{f|c} =  f_\varDelta(\alpha z_{f|c};\theta_\varDelta), \quad \alpha=\sigma\left(f_\alpha\left(\left[z_{f|c}||z_{r|c}\right] ;\theta_\alpha\right) \right)   ,
\end{equation}
where  $f_\alpha$ computes the similarity $\alpha$ between $z_{r|c}$ and $z_{f|c}$, parameterized by $\theta_\alpha$, 
while $f_\varDelta$ generates the offset  $\varDelta u_{f|c}^1$ based on $\alpha z_{f|c}$,  parameterized by $\theta_\varDelta$.
$\varDelta u_{f|c}$ quantifies the offset of the fragile rationale component relative to robust rationale component, 
and the center of fragile component is $u_{f|c}= u_{r|c} + \varDelta u_{f|c}$.  
Formally, the loss function of RFRM for $u_{r|c}$ is defined as $\mathcal{L}_{rrgm}^c$ directly based on Eq. (\ref{rrgm}), 
while the loss function $\mathcal{L}_{rfrm}^c$ for $u_{f|c}$ can be defined by reformulating Eq. (\ref{rfrm}), as follows:  
\begin{equation}
\small
  \begin{split}
  \hspace{-0.5cm}   \mathcal{L}_{rrgm}^c \hspace{-0.1cm}  & = \mathbb{E}_{\tilde{x} \sim p(\mathcal{X})} 
  \left[ \underset{y=0}{-\text{logsumexp}} \hspace{-0.1cm} \left(\hspace{-0.1cm} -\frac{\left\lVert\mathcal{F}_{\theta}(\tilde{x}|h_{r|c}) - u_{r|c}\right\rVert_2^2}{2} +   \epsilon_y \hspace{-0.1cm} \right) \right. \\
  \hspace{-0.4cm}   & \left. - \text{log}\left\lvert\text{det}J\right\rvert  + \frac{d_z}{2}\text{log}(2\pi)\right],   
  \end{split}
\label{eq131}
\end{equation}
\begin{equation}
\small
  \begin{split}
  \hspace{-0.5cm} & \mathcal{L}_{rfrm}^c  =  \mathbb{E}_{(\tilde{x},y) \sim p(\mathcal{X},Y)} 
  \left[ -\text{log}\left\lvert\text{det}J\right\rvert \right. \\ 
  \hspace{-0.5cm} &  \left. \hspace{-0.1cm} - \underset{y=0}{\text{logsumexp}} \hspace{-0.1cm}\left( \hspace{-0.1cm}
  -\frac{\left\lVert \mathcal{F}_{\theta}(\tilde{x}|h_{f|c}) \hspace{-0.1cm} -\hspace{-0.1cm}  \left(\cancel{\mathbf{G}}\hspace{-0.1cm} \left[u_{r|c}\right] \hspace{-0.1cm} + \hspace{-0.1cm} 
   \varDelta u_{f|c} \right) \right\rVert_2^2}{2} \hspace{-0.1cm} + \hspace{-0.05cm}  \epsilon_{y}^{k} \hspace{-0.1cm} \right) \hspace{-0.1cm} \right].
  \end{split}
\label{eq13}   
\end{equation}

For the abnormal class (i.e., $Y=1$), we employ a two-component Gaussian mixture model. 
The  class central  $u_o$  and the learnable offset $\varDelta u_o$  are  optimized by reformulating  Eq. (\ref{rrgm}) and 
Eq. (\ref{rfrm}). The corresponding loss functions $\mathcal{L}_{rrgm}^o$ and $\mathcal{L}_{rfrm}^o$  are defined as follows:
\begin{equation}
\small
  \begin{split}
  \hspace{-0.5cm}   \mathcal{L}_{rrgm}^o \hspace{-0.1cm}  & = \mathbb{E}_{\tilde{x} \sim p(\mathcal{X})} 
  \left[ \underset{y=1}{-\text{logsumexp}} \hspace{-0.1cm} \left(\hspace{-0.1cm} -\frac{\left\lVert\mathcal{F}_{\theta}(\tilde{x}|h_o) - u_o\right\rVert_2^2}{2} +   \epsilon_y \hspace{-0.1cm} \right) \right. \\
  \hspace{-0.5cm}   & \left. - \text{log}\left\lvert\text{det}J\right\rvert  + \frac{d_z}{2}\text{log}(2\pi)\right],   
  \end{split}
\end{equation}
\begin{equation}
\small
  \begin{split}
  \hspace{-0.5cm}   \mathcal{L}_{rfrm}^o & = \mathbb{E}_{(\tilde{x},y) \sim p(\mathcal{X},Y)} 
  \left[ -\text{log}\left\lvert\text{det}J\right\rvert \right. \\ 
  \hspace{-0.5cm} &  \left. \hspace{-0.1cm} - \underset{y=1}{\text{logsumexp}} \hspace{-0.1cm}\left( \hspace{-0.1cm}
  -\frac{\left\lVert \mathcal{F}_{\theta}(\tilde{x}|h_o) \hspace{-0.1cm} -\hspace{-0.1cm}  \left(\cancel{\mathbf{G}}\hspace{-0.1cm} \left[u_{o}\right] \hspace{-0.1cm} + \hspace{-0.1cm} 
  \varDelta u_{o} \right) \right\rVert_2^2}{2} \hspace{-0.1cm} + \hspace{-0.05cm}  \epsilon^k_{y} \hspace{-0.1cm}\right)\hspace{-0.1cm}\right].
  \end{split}
\end{equation}

Finally, the overall loss function of our RGMN model is defined as:
\begin{equation}
  \mathcal{L}_{rgmn} = \gamma (\mathcal{L}_{rrgm}^c + \mathcal{L}_{rfrm}^c)   + (1 - \gamma) (\mathcal{L}_{rrgm}^o + \mathcal{L}_{rfrm}^o),
\label{Lcgmn}
\end{equation}
where  hyperparameter $\gamma$  controls the balance between the rationale and non-rationale Gaussian mixture normalizing flow losses.
 
\subsection{The Overall Objective and Anomaly Scoring}
\hspace{1em}{\emph{a) Overall Objective.}
In the training phase, the overall objective of RagGAD is formulated as follows:
\begin{equation}
  \mathcal{L} =  \lambda_1  \mathcal{L}_{rgmn} + \lambda_2 \mathcal{L}_{rcon} + \lambda_3 \mathcal{L}_{spr}, 
\label{overalloss}
\end{equation}    
where $\mathcal{L}_{rcon} = \|\mathcal{X}-\hat{\mathcal{X}}\|_2^2$ is the attribute reconstruction loss. 
$\mathcal{L}_{spr}=|\mathcal{G}_c|$ is the rationale sparsity loss, 
which encourages RagGAD to automatically discard weak non-rationale correlations during the optimization process.
The hyperparameters $\lambda_1$, $\lambda_2$, and $\lambda_3$ control the relative contributions of each loss term and are tuned via grid search.

\hspace{1em}{\emph{b) Anomaly Scoring.}
Since anomalies deviate significantly from the majority of data instances, 
we hypothesize that their densities are low. 
Therefore, for each node $v_i$,  
based on Eq. (\ref{eq17}), the overall log-density conditioned on the rationale and non-rationale representations 
$h_{o}$, $h_{r|c}$, and $h_{f|c}$ is computed as follows: 
\begin{equation}
\small
    \text{log} p_{\mathcal{X}}(\tilde{x}|h)  
    \hspace{-0.05cm}  =   \hspace{-0.05cm}  \text{log} p_{\mathcal{X}}(\tilde{x}|h_o)
    \hspace{-0.05cm}  +   \hspace{-0.05cm}  \text{log} p_{\mathcal{X}}(\tilde{x}|h_{r|c})
    \hspace{-0.05cm}  +   \hspace{-0.05cm}  \text{log} p_{\mathcal{X}}(\tilde{x}|h_{f|c}).  
\end{equation}
During testing, the anomaly score $\mathcal{S}(x_i)$ for each node $v_i$ is defined as the negative log-density, 
i.e., $\mathcal{S}(x_i)= - \text{log} p_{\mathcal{X}}(\tilde{x}|h)$. 
A higher $\mathcal{S}(x_i)$ indicates a higher likelihood of $v_i$ being abnormal.

\begin{table*}
\vspace{-0.4cm}
\centering
\renewcommand{\arraystretch}{0.9} 
\caption{AUROC and AUPRC results (in percentage) on six real-world UGAD datasets with injected/real anomalies. 
Higher AUROC/AUPRC indicates better performance.
The best results are shown in \textcolor{deepgreen}{\textbf{bold}}, and the second-best results are  \textcolor{deepblue}{\textbf{bold}}.}  
\vspace{-0.3cm}
\resizebox{0.9\textwidth}{!}{
\begin{tabular}{c|r|cccccc} 
\toprule
\rowcolor[HTML]{F0F0F0}   &   \multicolumn{1}{c|}{}  &  \multicolumn{6}{c}{\textbf{Dataset}}    \\    
\cline{3-8} 
\rowcolor[HTML]{F0F0F0}  
\multirow{-2}{*}{\textbf{Metric}} & \multirow{-2}{*}{\quad\quad\textbf{Method}\quad\quad} 
& \textbf{BlogCatalog} & \textbf{ACM} & \textbf{Amazon} & \textbf{Facebook} & \textbf{Reddit} & \textbf{YelpChi}  \\ 
\midrule


& Anomalous\cite{peng2018anomalous} & $56.52_{\pm 2.50}$ & $68.56_{\pm 6.30}$ & $44.57_{\pm 0.30}$ & $90.21_{\pm 0.50}$ & $53.87_{\pm 1.20}$ & $49.56_{\pm 0.30}$ \\

& DOMINANT\cite{ding2019deep} & $75.90_{\pm 1.00}$ & $85.69_{\pm 2.00}$ & $59.96_{\pm 0.40}$ & $56.77_{\pm 0.20}$ & $55.55_{\pm 1.10}$ & $41.33_{\pm 1.00}$ \\

& CoLA\cite{liu2021anomaly} & $77.46_{\pm 0.90}$ & $82.33_{\pm 0.10}$ & $58.98_{\pm 0.80}$ & $84.34_{\pm 1.10}$ &  $60.28_{\pm 0.70}$ & $46.36_{\pm 0.10}$ \\

& SL-GAD\cite{zheng2021generative} & $81.23_{\pm 0.20}$ & $84.79_{\pm 0.50}$ & $59.37_{\pm 1.10}$ & $79.36_{\pm 0.50}$ & $56.77_{\pm 0.50}$ & $33.12_{\pm 3.50}$ \\

\textbf{AUROC} & HCM-A\cite{huang2022hop} & $79.80_{\pm 0.40}$ & $80.60_{\pm 0.40}$ & $39.56_{\pm 1.40}$ & $73.87_{\pm 3.20}$ & $45.93_{\pm 1.10}$ & $45.93_{\pm 0.50}$ \\

& ComGA\cite{luo2022comga} & $76.83_{\pm 0.40}$ & $82.21_{\pm 2.50}$ & $58.95_{\pm 0.80}$ & $60.55_{\pm 0.00}$ & $54.53_{\pm 0.30}$ & $43.91_{\pm 0.00}$ \\

& GADAM\cite{chen2024boosting} & $79.92_{\pm 0.00}$ &  \cellcolor{color2}$\textcolor{deepblue}{\textbf{94.66}_{\pm 0.00}}$ & $61.71_{\pm 0.00}$ & $94.03_{\pm 0.00}$ & $56.54_{\pm 0.00}$ & $41.77_{\pm 0.00}$ \\

& TAM\cite{qiao2024truncated} &  \cellcolor{color2}$\textcolor{deepblue}{\textbf{82.48}_{\pm 0.30}}$ & $88.78_{\pm 2.40}$ & $70.64_{\pm 1.00}$ & $91.44_{\pm 0.80}$ & $60.23_{\pm 0.40}$ & $56.43_{\pm 0.70}$ \\  


& HUGE\cite{pan2025label} & $62.13_{\pm 0.00}$ & $83.27_{\pm 0.00}$ & $85.16_{\pm 0.00}$ &  \cellcolor{color2}$\textcolor{deepblue}{\textbf{97.60}_{\pm 0.00}}$ & $59.06_{\pm 0.00}$ & $60.13_{\pm 0.00}$  \\

& CoCo\cite{wang2025context}  &  $79.10_{\pm 0.93}$ &   $88.37_{\pm 1.56}$ &   \cellcolor{color2}$\textcolor{deepblue}{\textbf{88.96}_{\pm 3.11}}$ &  $96.75_{\pm 0.01}$ &   \cellcolor{color2}$\textcolor{deepblue}{\textbf{61.92}_{\pm 0.51}}$  & $63.89_{\pm 0.06}$ \\ 

& SmoothGNN\cite{dong2025smoothgnn}   & --  & --  & $83.17_{\pm 0.88}$ &   $44.46_{\pm 1.08}$ &   $56.87_{\pm 1.69}$   &  $57.48_{\pm 0.06}$    \\

& GCTAM\cite{zhang2025gctam} & $62.47_{\pm 0.70}$ & $90.77_{\pm 1.50}$ & $84.38_{\pm 1.40}$ & $92.38_{\pm 0.30}$ & $59.21_{\pm 0.60}$ &  \cellcolor{color2}$\textcolor{deepblue}{\textbf{79.00}_{\pm 0.50}}$ \\
 
& FreeGAD\cite{zhao2025freegad} & $74.84_{\pm 0.00}$ &  $84.84_{\pm 0.00}$  &   $88.57_{\pm 0.00}$ &   $91.51_{\pm 0.00}$   &  $57.21_{\pm 0.00}$  & $78.55_{\pm 0.00}$ \\

\cdashline{3-8}
& \multicolumn{1}{c|}{RagGAD} & \cellcolor{color1}$\textcolor{deepgreen}{\textbf{87.32}_{\pm 0.67}}$ & \cellcolor{color1}$\textcolor{deepgreen}{\textbf{97.24}_{\pm 0.34}}$ & \cellcolor{color1}$\textcolor{deepgreen}{\textbf{92.04}_{\pm 0.21}}$ & \cellcolor{color1}$\textcolor{deepgreen}{\textbf{98.62}_{\pm 0.27}}$ & \cellcolor{color1}$\textcolor{deepgreen}{\textbf{63.71}_{\pm 0.71}}$ & \cellcolor{color1}$\textcolor{deepgreen}{\textbf{81.13}_{\pm 0.14}}$  \\  
\midrule


& Anomalous\cite{peng2018anomalous} & $6.52_{\pm 0.50}$ & $6.35_{\pm 0.60}$ & $5.58_{\pm 0.10}$ & $18.98_{\pm 0.40}$ & $3.75_{\pm 0.40}$ & $5.19_{\pm 0.20}$ \\

& DOMINANT\cite{ding2019deep} & $31.02_{\pm 1.10}$ & $44.02_{\pm 3.60}$ & $14.24_{\pm 0.20}$ & $3.14_{\pm 4.10}$ & $3.56_{\pm 0.20}$ & $3.95_{\pm 2.00}$ \\

& CoLA\cite{liu2021anomaly} & $32.70_{\pm 0.00}$ & $32.35_{\pm 1.70}$ & $6.77_{\pm 0.10}$ & $21.06_{\pm 1.70}$ & $4.49_{\pm 0.20}$ & $4.48_{\pm 0.20}$  \\

& SL-GAD\cite{zheng2021generative} & $38.82_{\pm 0.70}$ & $37.84_{\pm 1.10}$ & $6.34_{\pm 0.50}$ & $13.16_{\pm 2.00}$ & $4.06_{\pm 0.40}$ & $3.50_{\pm 0.00}$ \\

\textbf{AUPRC} & HCM-A\cite{huang2022hop} & $31.39_{\pm 0.10}$ & $34.13_{\pm 0.40}$ & $5.27_{\pm 1.50}$ & $7.13_{\pm 0.40}$ & $2.87_{\pm 0.50}$ & $2.87_{\pm 1.20}$ \\

& ComGA\cite{luo2022comga} & $32.93_{\pm 2.80}$ & $28.73_{\pm 1.20}$ & $11.53_{\pm 0.50}$ & $3.54_{\pm 0.10}$ & $3.74_{\pm 0.10}$ & $4.23_{\pm 0.00}$  \\

& GADAM\cite{chen2024boosting} & $28.14_{\pm 0.00}$ & $33.77_{\pm 0.00}$ & $11.74_{\pm 0.00}$ & $24.06_{\pm 0.00}$ & $4.62_{\pm 0.00}$ & $4.23_{\pm 0.00}$ \\ 

& TAM\cite{qiao2024truncated} & \cellcolor{color2}$\textcolor{deepblue}{\textbf{41.82}_{\pm 0.50}}$ & $51.24_{\pm 1.80}$ & $26.34_{\pm 0.80}$ & $22.33_{\pm 1.60}$ & $4.46_{\pm 0.10}$ &  $7.78_{\pm 0.90}$ \\  


& HUGE\cite{pan2025label} & $7.44_{\pm 0.00}$ & $32.93_{\pm 0.00}$ & $66.72_{\pm 0.00}$ &  \cellcolor{color2}$\textcolor{deepblue}{\textbf{36.74}_{\pm 0.00}}$ &  \cellcolor{color2}$\textcolor{deepblue}{\textbf{5.11}_{\pm 0.00}}$ & $7.08_{\pm 0.00}$   \\

& CoCo\cite{wang2025context}  &  $37.53_{\pm 0.36}$   &   $40.19_{\pm 2.44}$ &  $39.20_{\pm 0.06}$  &  $36.31_{\pm 0.08}$ &  $5.02_{\pm 0.17}$ &   $7.18_{\pm 0.01}$\\ 

& SmoothGNN\cite{dong2025smoothgnn}   & --  & --  & $39.05_{\pm 2.33}$ &   $1.95_{\pm 0.04}$ &   $4.17_{\pm 0.16}$   &   \cellcolor{color2}$\textcolor{deepblue}{\textbf{18.18}_{\pm 0.03}}$  \\

& GCTAM\cite{zhang2025gctam} & $12.90_{\pm 0.65}$ &  \cellcolor{color2}$\textcolor{deepblue}{\textbf{52.10}_{\pm 0.70}}$ & $50.69_{\pm 8.10}$ & $22.81_{\pm 0.60}$ & $4.17_{\pm 0.10}$  &   $16.04_{\pm 1.00}$ \\
 
& FreeGAD\cite{zhao2025freegad} &  $34.03_{\pm 0.00}$ &  $37.15_{\pm 0.00}$  &   \cellcolor{color2}$\textcolor{deepblue}{\textbf{75.06}_{\pm 0.00}}$ &   $22.50_{\pm 0.00}$   &  $3.85_{\pm 0.00}$  &   $15.80_{\pm 0.00}$  \\

\cdashline{3-8}
& \multicolumn{1}{c|}{RagGAD} & \cellcolor{color1}$\textcolor{deepgreen}{\textbf{46.32}_{\pm 1.21}}$ &  \cellcolor{color1}$\textcolor{deepgreen}{\textbf{55.27}_{\pm 0.21}}$ & \cellcolor{color1}$\textcolor{deepgreen}{\textbf{77.67}_{\pm 0.11}}$ & \cellcolor{color1}$\textcolor{deepgreen}{\textbf{39.33}_{\pm 0.42}}$ & \cellcolor{color1}$\textcolor{deepgreen}{\textbf{5.45}_{\pm 0.20}}$ & \cellcolor{color1}$\textcolor{deepgreen}{\textbf{19.38}_{\pm 0.43}}$  \\  
\bottomrule
\end{tabular}}
\label{result_s} 
\vspace{-0.3cm}
\end{table*}

\section{Experiments}
\subsection{Experimental Setup}
\hspace{1em}{\emph{a) Datasets.}
We evaluate  RagGAD on ten benchmark datasets for UGAD,  following TAM \cite{qiao2024truncated} and HUGE \cite{pan2025label}. 
These datasets comprise six  real-world datasets, 
including  BlogCatalog \cite{tang2009relational}, ACM \cite{tang2008arnetminer},  Amazon \cite{dou2020enhancing}, 
Facebook \cite{xu2022contrastive}, Reddit \cite{kumar2019predicting}, and YelpChi \cite{kumar2019predicting}.
Additionally, four large-scale datasets are used, namely  Amazon-all \cite{qiao2024truncated},  YelpChi-all \cite{chen2022gccad},  
T-Finance \cite{tang2022rethinking}, and OGB-Proteins \cite{hu2020open}.
Among these datasets, BlogCatalog, ACM, and OGB-Protein contain two types of injected anomalies, contextual and structural anomalies.
Following \cite{qiao2024truncated,pan2025label},  all graphs are converted to homogeneous undirected graphs for consistency during evaluation.
 
\hspace{1em}{\emph{b) Baselines.}
We compare RagGAD with a comprehensive set of state-of-the-art baselines, 
including 
Anomalous \cite{peng2018anomalous}, 
DOMINANT \cite{ding2019deep}, CoLA \cite{liu2021anomaly},   SL-GAD \cite{zheng2021generative},   HCM-A \cite{huang2022hop}, 
ComGA \cite{luo2022comga},  GADAM \cite{chen2024boosting}, TAM \cite{qiao2024truncated}, HUGE \cite{pan2025label},
SmoothGNN \cite{dong2025smoothgnn}, GCTAM \cite{zhang2025gctam}, CoCo \cite{wang2025context}, and  FreeGAD \cite{zhao2025freegad}. 
Following \cite{qiao2024truncated,liu2024arc}, two evaluation metrics,
Area Under the Receiver Operating Characteristic Curve (AUROC) 
and Area Under the precision recall curve (AUPRC), are used for UGAD. 
The reported AUROC and AUPRC results are averaged over five independent runs with different random seeds.
Reported performance for all baselines is obtained from publicly available results or reproduced using their official source code.

\hspace{1em}{\emph{c) Implementation Details.}
\label{Implementation Details}
RagGAD adopts a consistent strategy for hyperparameter tuning, early stopping, and model selection, 
following TAM \cite{qiao2024truncated} and HUGE \cite{pan2025label}.  
All experiments are performed using the PyTorch framework on an NVIDIA GeForce RTX 3090 (24GB) GPU.
The model is optimized with the Adam optimizer over 500 epochs with a learning rate of 1e-4. 
The projector $\mathcal{P}(\cdot)$ employs the PCA method, 
while the graph convolution layer $\text{GCN}(\cdot)$ is implemented as a two-layer GCN. 
The reconstructor $\mathcal{R}(\cdot)$ is designed as a two-layer MLP to reconstruct node attributes.

\subsection{Results}
\hspace{1em}{\emph{a) Comparison with Baselines.}
Table \ref{result_s} presents the AUROC and AUPRC results for all baselines
 across six real-world UGAD datasets with both real and injected anomalies. 
Overall, RagGAD consistently achieves the best performance across all datasets under both metrics, 
demonstrating its strong effectiveness and robustness.
In terms of AUPRC, which is more sensitive to class imbalance, RagGAD also achieves the best performance across all datasets. 
RagGAD surpasses TAM on BlogCatalog and ACM by +4.50\% (46.32 vs. 41.82) and +3.17\% (55.27 vs. 52.10), respectively. 
On Amazon, RagGAD improves over FreeGAD by +2.61\% (77.67 vs. 75.06), 
while on Facebook, it exceeds HUGE by +2.59\% (39.33 vs. 36.74). 
Although the absolute gains on Reddit and YelpChi are relatively smaller (+0.34\% and +1.20\%), 
RagGAD still consistently ranks first, demonstrating stable performance under highly imbalanced and noisy conditions.
In terms of AUROC, RagGAD ranks first on all six datasets, outperforming the second-best methods by clear margins. 
Specifically, it improves over TAM on BlogCatalog by +4.84\% (87.32 vs. 82.48), 
over GADAM on ACM by +2.58\% (97.24 vs. 94.66), and over CoCo on Amazon by +3.08\% (92.04 vs. 88.96). 
On Facebook, RagGAD achieves 98.62\%, exceeding the strong baseline HUGE (97.60) by +1.02\%. 
Even on more challenging datasets such as Reddit and YelpChi, 
RagGAD still attains the highest scores (63.71\% and 81.13\%), 
surpassing the second-best methods by +1.79\% and +2.13\%, respectively. 
Moreover, compared with recent strong methods such as HUGE, TAM, and FreeGAD, 
which exhibit dataset-dependent performance variations, RagGAD maintains consistently high performance across all scenarios. 

\begin{table} 
\centering
\footnotesize 
\renewcommand{\arraystretch}{0.8} 
\setlength{\tabcolsep}{0.03cm}
\caption{Results on large-scale graphs demonstrate the effectiveness of each baseline for handling large numbers of nodes and edges. 
$\text{OOM}$ indicates out-of-memory on a 24GB GPU.}
\vspace{-0.3cm}
\resizebox{0.49\textwidth}{!}{
\begin{tabular}{c|r|cccc}
\toprule
\rowcolor[HTML]{F0F0F0} &  &  \multicolumn{4}{c}{{\textbf{Dataset}}}  \\ 
\cdashline{3-5}
\rowcolor[HTML]{F0F0F0}  
\multirow{-2}{*}{{\textbf{Metric}}} 
& \multirow{-2}{*}{\quad \quad \textbf{Method} \quad \quad}      
& {\textbf{Amazon-all}} 
& {\textbf{YelpChi-all}} 
& {\textbf{T-Finance}} 
& {\textbf{OGB-Proteins}}  \\ 
\midrule

& DOMINANT\cite{ding2019deep}  & 69.37 & 53.90 & 53.80 & 72.67  \\
& ComGA\cite{luo2022comga}     & 71.54 & 53.52 & 55.42 & 71.34  \\
& CoLA\cite{liu2021anomaly} & 26.14 & 48.01 & 48.29 & 71.42  \\  
\textbf{AUROC}  
& SL-GAD\cite{zheng2021generative}    & 27.28 & 55.51 & 46.48 & 73.71  \\
& GADAM\cite{chen2024boosting}     & 41.55 & 47.95 & 25.38 & 73.38   \\
& TAM\cite{qiao2024truncated}       & 84.76 & 58.18 & 61.75 &  \cellcolor{color2}\textcolor{deepblue}{\textbf{74.49}}  \\  
& GCTAM\cite{zhang2025gctam}    & 88.89 &  \cellcolor{color2}\textcolor{deepblue}{\textbf{59.57}} &  $\text{OOM}$  &   $\text{OOM}$   \\
& CoCo\cite{wang2025context}   & 85.61  & OOM & OOM & OOM \\
& HUGE\cite{pan2025label}      &  \cellcolor{color2}\textcolor{deepblue}{\textbf{88.92}} & 57.67 & $\text{OOM}$  & $\text{OOM}$  \\   
& FreeGAD\cite{zhao2025freegad} & 87.83  & 54.17 &   \cellcolor{color2}\textcolor{deepblue}{\textbf{92.13}} &  66.51 \\
\cdashline{3-6}
& \multicolumn{1}{c|}{RagGAD}   &  \cellcolor{color1}\textcolor{deepgreen}{\textbf{93.15}} &  \cellcolor{color1}\textcolor{deepgreen}{\textbf{63.12}} &  \cellcolor{color1}\textcolor{deepgreen}{\textbf{92.98}} &  \cellcolor{color1}\textcolor{deepgreen}{\textbf{78.83}}  \\ 
\midrule

& DOMINANT\cite{ding2019deep}  & 10.15 & 16.38 & 4.74 & 22.17 \\
& ComGA\cite{luo2022comga}     & 18.54 & 16.58 & 4.81 & 15.54 \\
& CoLA\cite{liu2021anomaly}      & 5.16 & 13.61 & 4.10 & 13.49  \\
& GADAM\cite{chen2024boosting}     & 5.78 & 13.79 & 3.04 & 21.04  \\
\textbf{AUPRC}   
& SL-GAD\cite{zheng2021generative}    & 4.44 & 17.11 & 3.86 & 17.71  \\
& TAM\cite{qiao2024truncated}      & 43.46 & 18.86 & 5.47 & 21.73 \\
& GCTAM\cite{zhang2025gctam}     & 67.18 &  \cellcolor{color2}\textcolor{deepblue}{\textbf{20.13}} &  $\text{OOM}$  &   $\text{OOM}$  \\
& CoCo\cite{wang2025context}   & 70.36 & OOM & OOM & OOM\\
& HUGE\cite{pan2025label}      & \cellcolor{color2}\textcolor{deepblue}{\textbf{76.68}} & 18.69 & $\text{OOM}$  & $\text{OOM}$  \\
& FreeGAD\cite{zhao2025freegad} & 66.32 & 16.96  &    \cellcolor{color2}\textcolor{deepblue}{\textbf{73.71}} &   \cellcolor{color2}\textcolor{deepblue}{\textbf{25.21}}  \\
\cdashline{3-6}
& \multicolumn{1}{c|}{RagGAD}    &  \cellcolor{color1}\textcolor{deepgreen}{\textbf{80.02}} &   \cellcolor{color1}\textcolor{deepgreen}{\textbf{23.25}} &  \cellcolor{color1}\textcolor{deepgreen}{\textbf{74.81}} &  \cellcolor{color1}\textcolor{deepgreen}{\textbf{31.44}} \\ 
\bottomrule
\end{tabular}}
\label{result_l} 
\vspace{-0.4cm}
\end{table}

\begin{table*}[t] 
\vspace{-0.4cm}
\centering
\footnotesize 
\renewcommand{\arraystretch}{0.85} 
\setlength{\tabcolsep}{0.1cm}
\caption{Ablation studies for key components in RagGAD (in percentage). 
We report the average \textbf{AUROC} and \textbf{AUPRC} over five independent random runs.
\textcolor{deepred}{\textbf{IMP}}: the average improvement of each variant over the rest (in percentage).}
\vspace{-0.3cm}
\begin{tabular}{c|cc|cc|cc|cc|cc|cc} 
\toprule
\rowcolor[HTML]{F0F0F0}    
& \multicolumn{2}{c|}{\textbf{Amazon}} 
& \multicolumn{2}{c|}{\textbf{Facebook}} 
& \multicolumn{2}{c|}{\textbf{Reddit}}  
& \multicolumn{2}{c|}{\textbf{YelpChi-all}}  
& \multicolumn{2}{c|}{\textbf{T-Finance}}  
& \multicolumn{2}{c}{\textbf{Avg \textcolor{deepred}{(IMP)}}} \\  
\cdashline{1-13} 
\rowcolor[HTML]{F0F0F0}  
\multirow{-2}{*}{\textbf{Variant}} 
& \textbf{AUROC} & \textbf{AUPRC}  
& \textbf{AUROC} & \textbf{AUPRC}  
& \textbf{AUROC} & \textbf{AUPRC}  
& \textbf{AUROC} & \textbf{AUPRC}  
& \textbf{AUROC} & \textbf{AUPRC}  
& \textbf{AUROC} & \textbf{AUPRC}  \\  
\midrule

w/o ARD   
& 75.49 & 61.35 
& 93.15 & 19.88 
& 58.01 & 4.27 
& 54.52 & 17.84 
& 78.81 & 66.08 
& 72.00 \textcolor{deepred}{(0.00)}  & 33.88 \textcolor{deepred}{(0.00)}   \\

\rowcolor[HTML]{F0F0F0}  
w/o RGMN  
& 73.23 & 60.07 
& 95.36 & 22.68 
& 59.36 & 4.30 
& 55.68 & 18.09 
& 77.80 & 66.02 
& 72.29 \textcolor{deepred}{(0.29)} & 34.23 \textcolor{deepred}{(0.35)} \\

w/o RRGM  
& 86.53 & 66.88 
& 91.08 & 23.03 
& 60.37 & 4.58 
& 57.21 & 19.60 
& 84.23 & 66.79 
& 75.88 \textcolor{deepred}{(3.60)} & 35.82 \textcolor{deepred}{(1.94)} \\

\rowcolor[HTML]{F0F0F0}  
w/o RFRM  
& 90.58 & 76.03 
& 95.85 & 33.03 
& 61.04 & 4.80 
& 61.57 & 21.94 
& 90.84 & 70.23 
& 79.97 \textcolor{deepred}{(4.09)} & 38.91 \textcolor{deepred}{(5.03)} \\

\cdashline{1-13}

RagGAD      
&  92.04 & 77.67 &  98.62 & 39.33 & 63.71 & 5.45 & 63.12 & 23.25 &  92.98 &  74.81
&  82.09 \textcolor{deepred}{(2.21)} & 44.10 \textcolor{deepred}{(2.90)} \\    

\bottomrule
\end{tabular}
\label{keycomponts} 
\vspace{-0.3cm}
\end{table*}

\begin{figure*}[t]
\centering
\includegraphics[width=0.95\textwidth,height=0.4\textheight]{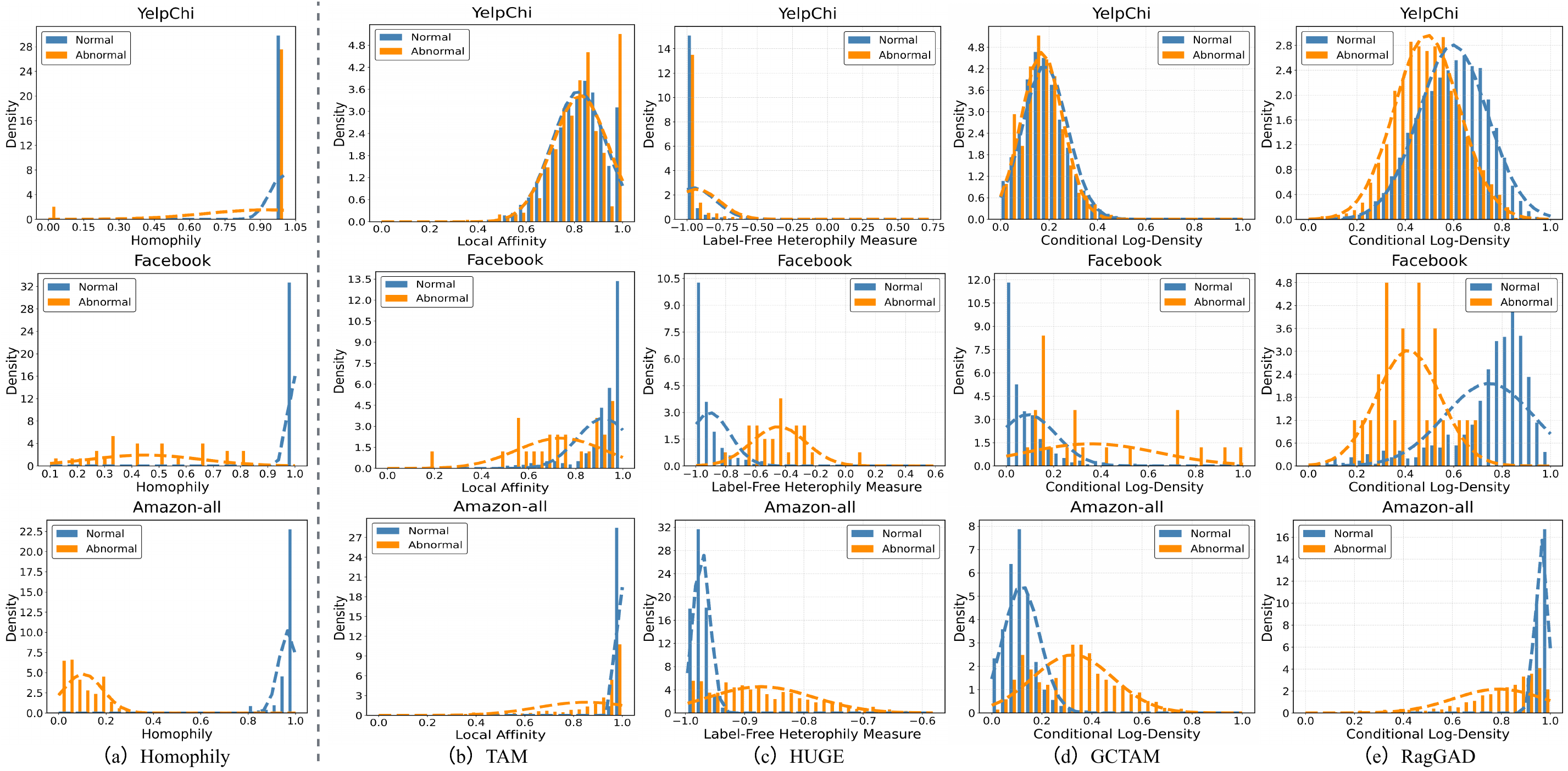} 
\vspace{-0.4cm}
\caption{Visualization of (a) homophily distribution of normal and abnormal nodes, and (b)-(e) the anomaly score of 
TAM (local affinity) \cite{qiao2024truncated}, HUGE (label-free heterophily measure) \cite{pan2025label}, 
GCTAM (contextual and global affinity) \cite{zhang2025gctam}, and RagGAD (rationale-aware conditional log-density) on YelpChi \cite{kumar2019predicting}, 
Facebook \cite{xu2022contrastive}, and Amazon-all \cite{qiao2024truncated} datasets. }   
\label{homophily}
\end{figure*}

\begin{figure}[t]
\vspace{-0.4cm}
\centering
\includegraphics[width=0.49\textwidth]{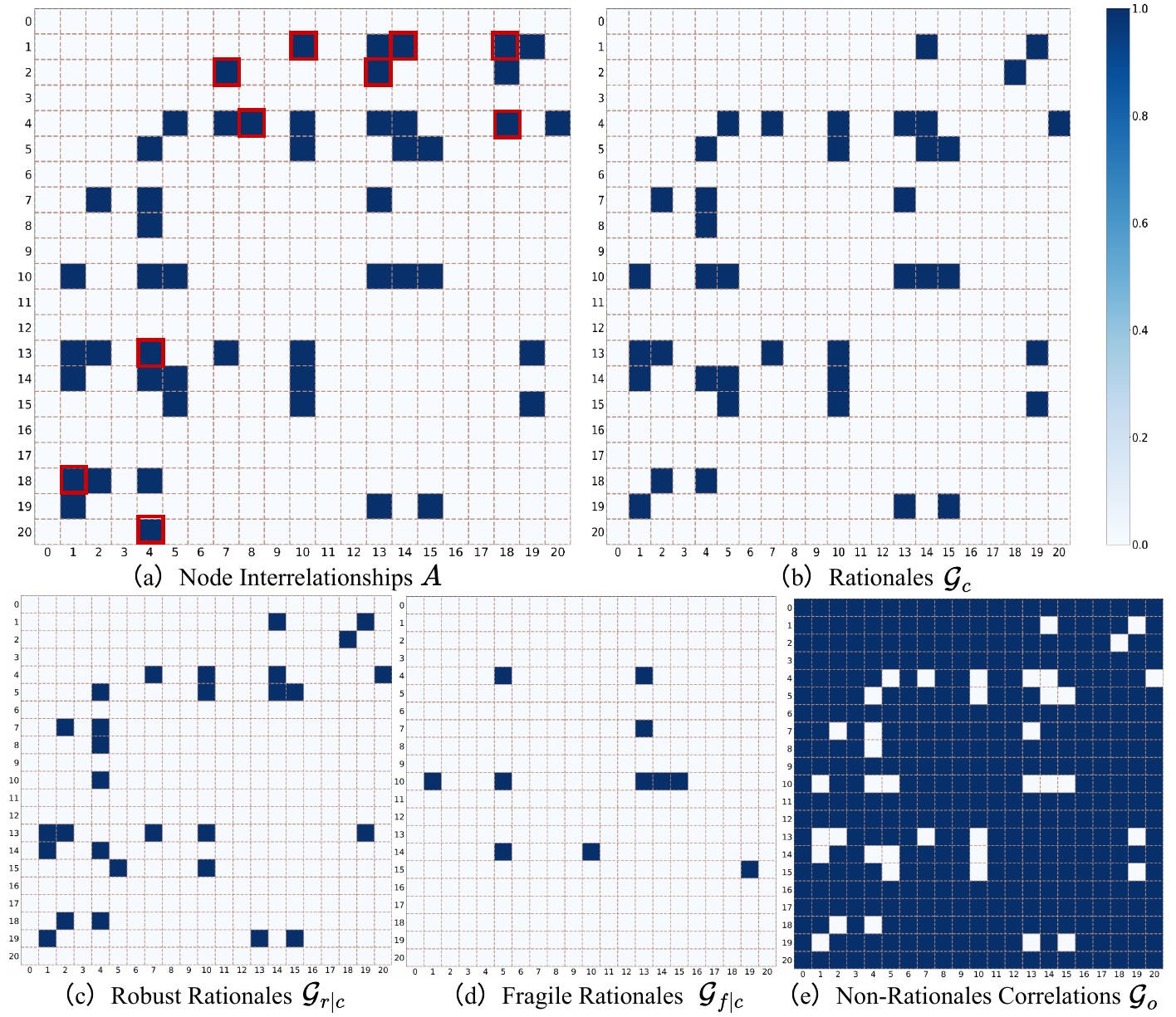} 
\vspace{-0.7cm}
\caption{The visualization of inter-node interrelationships, disentangled  non-rationale correlations and 
rationales (robust and fragile) relationships by RagGAD. 
The deeper the color, the stronger the relationship between nodes.  
The red box \textcolor{deepred}{\textbf{$\Box$}} indicates correlations between nodes  but are not rationales and can thus be removed.}
\label{graphs}
\end{figure}

\begin{figure}[t] 
\centering
\includegraphics[width=0.49\textwidth]{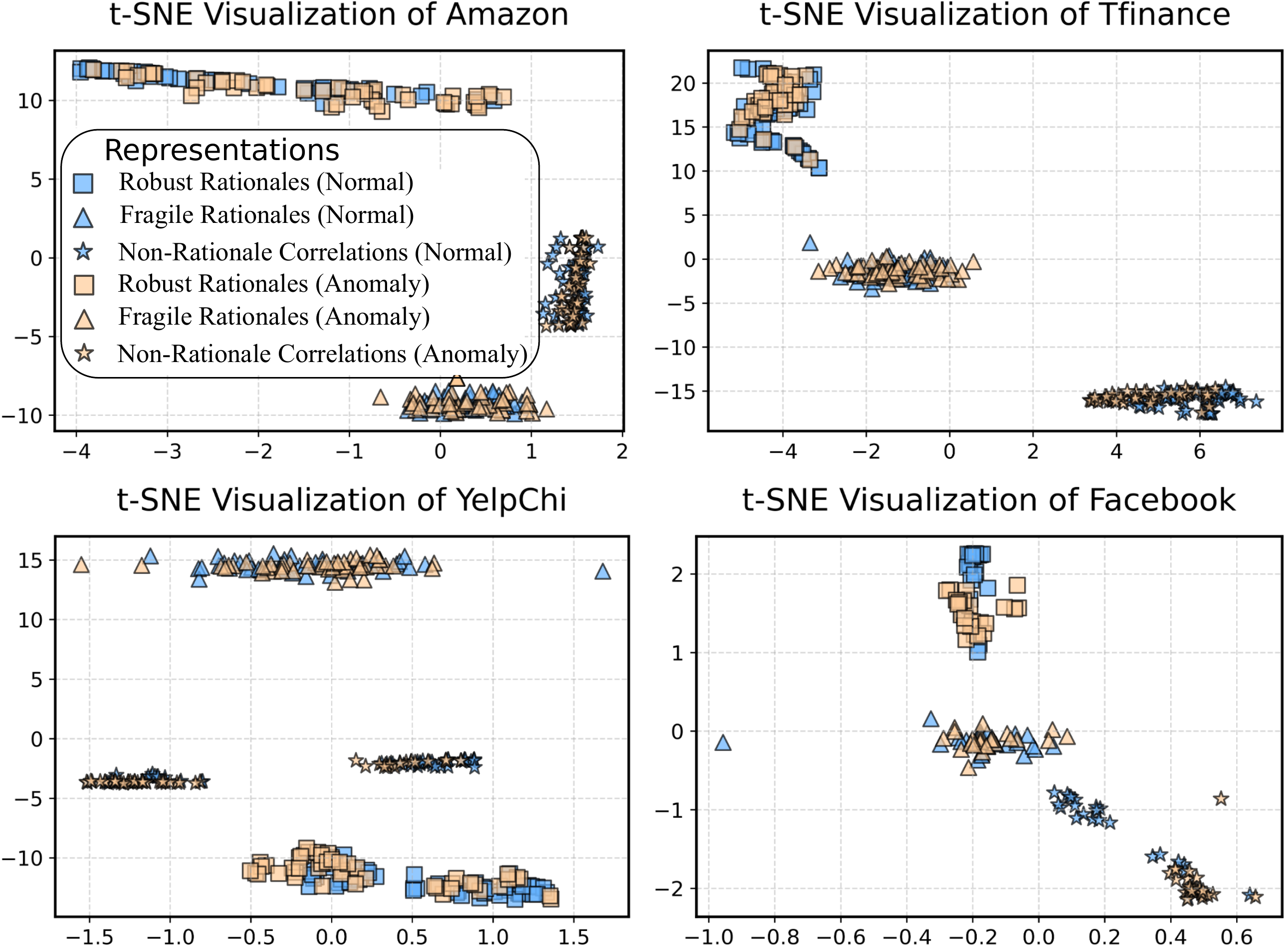} 
\vspace{-0.7cm}
\caption{The t-SNE visualization shows the robust rationale representation ($\blacksquare$), 
fragile rationale representation ($\blacktriangle$), and 
non-rationale representation ($\bigstar$) learned on four benchmark datasets.}
\label{tsne}
\end{figure}

\hspace{1em}{\emph{b) Performance on Large-scale Graphs.}    
We evaluated RagGAD on four large-scale graph datasets with substantial node and 
edge counts to assess its effectiveness in handling complex graph structures.
The results in Table  \ref{result_l} demonstrate  that RagGAD maintains high effectiveness despite the increased complexity of anomalies,
delivering consistent and significant performance across all four datasets.
In terms of AUROC, RagGAD outperforms the strongest baselines on all datasets, achieving 93.15\% on Amazon-all (+4.23\% over HUGE), 
63.12\% on YelpChi-all (+3.55\% over GCTAM), 92.98\% on T-Finance (+0.85\% over FreeGAD), and 78.83\% on OGB-Proteins (+4.34\% over TAM). 
On more challenging datasets such as T-Finance and OGB-Proteins, 
where anomalies are highly subtle and distributions are more complex, RagGAD still maintains clear advantages, 
indicating its robustness in capturing intricate structural patterns.
Consistent improvements are also observed under AUPRC, with notable gains such as +3.34\% on Amazon-all and +6.23\% on OGB-Proteins,
demonstrating its effectiveness under highly imbalanced settings. 
Moreover, several recent methods encounter out-of-memory (OOM) issues on larger datasets, 
whereas RagGAD remains stable and achieves strong performance across all settings.
These results demonstrate that modeling stable rationales while mitigating spurious correlations enables RagGAD to better generalize 
across diverse graph structures and anomaly types, leading to superior performance in large-scale graph anomaly detection.

\hspace{1em}{\emph{c) Visualization of Anomaly Score.}
Figure \ref{homophily}(a), presents the homophily distribution of normal and abnormal nodes across benchmark datasets.
The results indicate that the Facebook and Amazon-all datasets exhibit one-class homophily, 
characterized by stronger connectivity or affinity among normal nodes compared to abnormal nodes. 
In contrast, the YelpChi dataset deviates from this assumption, with connectivity or affinity 
among abnormal nodes being comparable to, or even surpassing, that among normal nodes.
Figure \ref{homophily}(b)-(e), depict the anomaly score distributions for TAM, HUGE, GCTAM, and RagGAD. 
As shown in Figure \ref{homophily}(b)-(e),  TAM fails to establish a clear boundary 
between the distributions of normal and abnormal nodes across all datasets. 
HUGE demonstrates improved performance on the Facebook and Amazon-all datasets,    
yet does not differentiate normal and abnormal nodes on the Facebook dataset. 
Although HUGE identifies the homophily trap and introduces a label-free heterophily measure strategy, 
it falls short of substantially enhancing the distributional distinction between normal and abnormal nodes. 
GCTAM extends TAM by incorporating contextual and global affinity to filter anomalous nodes,
but it still fails to effectively distinguish between normal and anomalous distributions on the YelpChi dataset.
Conversely, RagGAD achieves a clear distributional distinction between normal and abnormal 
nodes across all datasets. While some overlap in distribution density occurs on the YelpChi dataset, 
the anomaly score distributions of normal and abnormal nodes remain separated, 
exhibiting clearly differentiated peaks.

\hspace{1em}{\emph{d) Visualization of Rationale Graphs.}
In disentangling non-rationale correlations and rationales between nodes,
we visualize the disentangled non-rationale correlations $\mathcal{G}_o$ and rationales ($\mathcal{G}_{r|c}$ and $\mathcal{G}_{f|c}$) 
on the Facebook dataset, as shown in Figure \ref{graphs}. 
The highly challenging Facebook dataset, has the lowest anomaly ratio (2.49$\%$) 
and the smallest number of nodes, making it an ideal choice for this validation.
To improve visualization clarity, we randomly extract 20$\times$20 subgraphs from the 1,080-node of Facebook for display.
As shown in Figure \ref{graphs}(b), the disentangled rationales $\mathcal{G}_c$ 
are sparse, yet they retain most of the node correlations while eliminating certain spurious correlations (highlighted by the red box) from $A$.
Figure \ref{graphs}(c)-(e) demonstrate that robust rationales $\mathcal{G}_{r|c}$ dominate the graph, 
accounting for approximately 80$\%$ of the relationships, compared to fragile rationales $\mathcal{G}_{f|c}$.
However, despite their lower proportion, fragile rationales are crucial for diversifying normal patterns. 
When extended to the entire graph (1081 nodes), the number of fragile rationales becomes non-negligible. 
Essentially, the presence of fragile rationales greatly expands the diversity of normal node patterns. 
Therefore, as illustrated in Figure \ref{overview}, RagGAD further decomposes the rationales into 
robust and fragile components and performs robust-fragile rationale mixture learning to enhance generalization for anomaly detection.

\hspace{1em}{\emph{e) Visualization of Representation.}
Figure \ref{tsne} illustrates the t-SNE visualization of robust rationale, 
fragile rationale, and non-rationale representations extracted by RagGAD from benchmark datasets. 
The visualization demonstrates that RagGAD effectively disentangles rationale and non-rationale representations, 
further decomposing the rationale components into robust and fragile components. 
Within RagGAD, the adaptive rationale disentangler (ARD)  disentangles rationale and non-rationale 
correlations from node interdependencies, subsequently decomposing rationale into robust and 
fragile components. Leveraging rationales, RagGAD derives robust and fragile rationale representations, 
while non-rationale representations extracted from non-rationale correlations.
This enablesprecise modeling and discrimination of conditional densities for normal and abnormal nodes.
As shown in Figure \ref{tsne}, RagGAD achieves a compact intra-class feature distribution 
while distinctly separating clusters of different class representations.

\subsection{Ablation Study}
\hspace{1em}{\emph{a) Components Analysis.}   
We investigate the effectiveness of each component in RagGAD across five datasets, as detailed in Table \ref{keycomponts}.  
Table \ref{keycomponts} demonstrates a progressive improvement in accuracy as more components are integrated,
highlighting the essential role of each component in the overall effectiveness of RagGAD.
To evaluate the effectiveness of our ARD (w/o ARD), 
we remove rationale disentangler and instead utilize node interrelationships $\mathcal{I}$ 
to guide information propagation between nodes, performing density estimation based on conventional normalizing flow.
The results in Table \ref{keycomponts} indicate that compared to w/o RGMN, which incorporates 
ARD followed by a conventional normalizing flow, ARD improves model performance, further validating its effectiveness.
 
\begin{figure}[t]
\vspace{-0.4cm}
\centering
\includegraphics[width=0.35\textwidth,height=0.33\textheight]{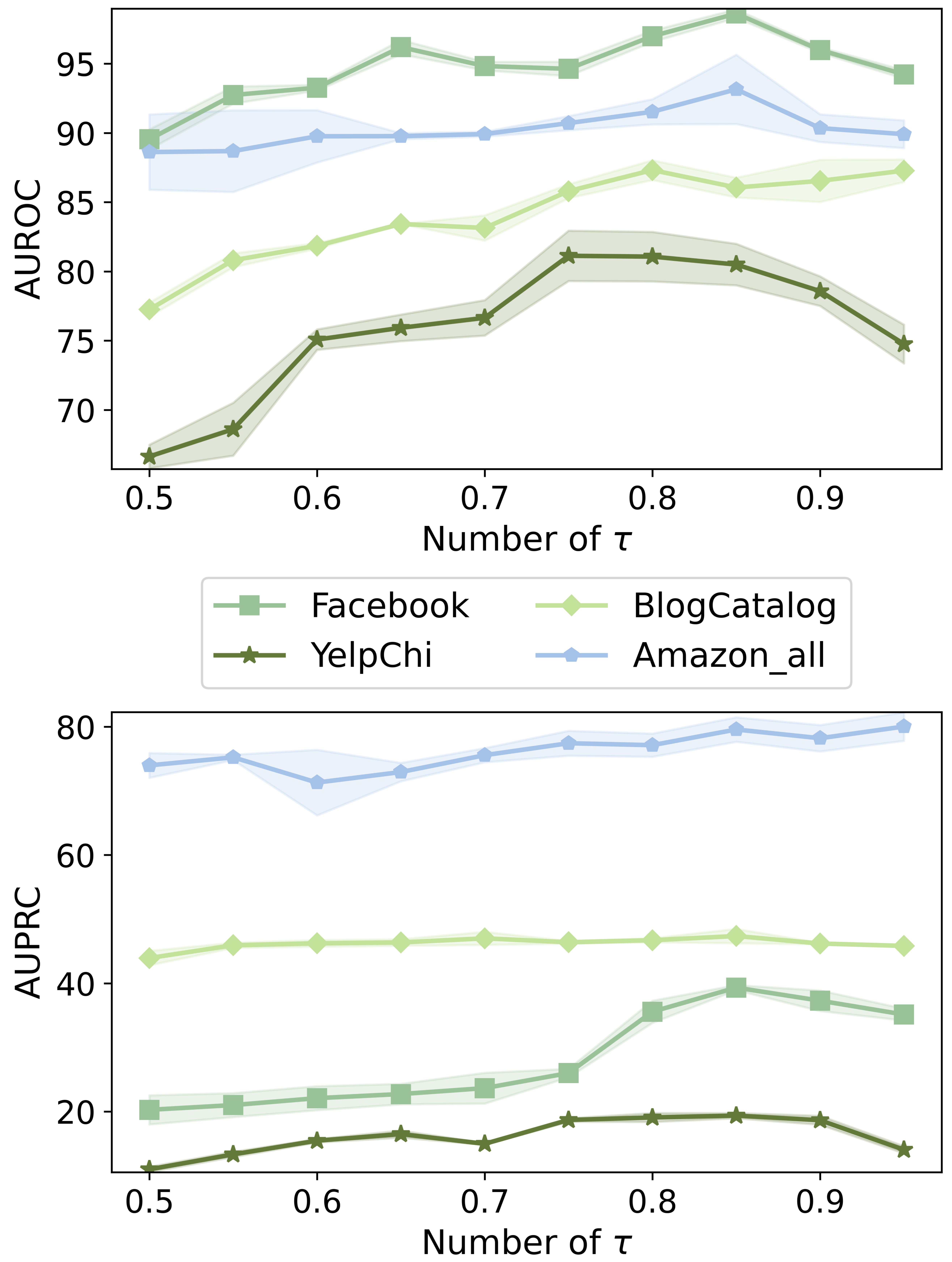} 
\vspace{-0.35cm}
\caption{AUROC and AUPRC of RagGAD w.r.t hyperparameter $\tau$.}   
\label{tau}
\end{figure}

We compare RagGAD with two variants that exclude the corresponding components, i.e., rationale-non-rationale 
gaussian mixture modeling (w/o RRGM) 
and robust-fragile rationale mixture learning (w/o RFRM), to evaluate the effectiveness of these key components. 
The results in the lower part of Table \ref{keycomponts} confirm that integrating RRGM and RFRM significantly improves 
model performance, with RRGM alone demonstrating strong competitiveness.
Notably, RRGM plays a pivotal role, as its removal results in a substantial drop in AUPRC and AUROC by approximately
1.94\% (35.82 vs. 33.88) and 3.88\% (75.88 vs. 72.00) across all datasets compared to RagGAD.
 
\begin{figure}[t]
\centering
\includegraphics[width=0.44\textwidth]{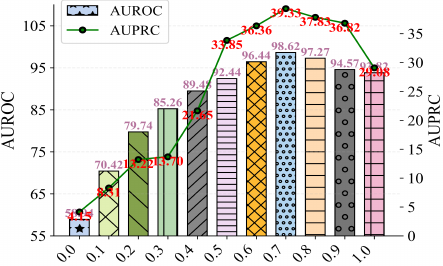} 
\vspace{-0.4cm}
\caption{Sensitivity analysis of hyperparameter  $\gamma$ on the Facebook dataset.  
}   
\label{gamma}
\vspace{-0.2cm}
\end{figure}

\hspace{1em}{\emph{b) Sensitivity Analysis of $\tau$.}
In order to evaluate the sensitivity of RagGAD to the hyperparameter $\tau$,  
we vary its value within the range [0.5, 1] in increments of 0.05 for validation.
The parameter $\tau$, which governs the selection of edges as rationales (see Eq.(\ref{causal_d})), is a crucial hyperparameter for RagGAD. 
As shown in Figure \ref{tau}, as $\tau$ increases, RagGAD exhibits a tred of initially rising, then flattening, and finally gradually decreaseing. 
This variation pattern is reasonable, as a very low value of $\tau$ forces the model to incorrectly classify 
non-rationale correlations as rationales, 
resulting in excessively high false positives.
In contrast, as $\tau$ increases, the rationale edge selection strategy becomes stricter. 
When $\tau$ is too high, the model may miss relatively fragile rationales, leading to an increase in rationale false negatives.
For AUROC, the impact of changing $\tau$ is relatively significant, with a noticeable performance growth between 0.5 and 0.8.
Whether for AUROC or AUPRC, when $\tau$ is around 0.85, the model achieves its best performance across all datasets. 
Therefore, in all our experiments, we set the default value of $\tau$ to 0.85.

\hspace{1em}{\emph{c) Sensitivity Analysis of $\gamma$.}     
The hyperparameter $\gamma$ controls the contributions of the rationales and non-rationale gaussian mixture normalizing flow loss terms (see Eq.(\ref{Lcgmn})).
We  adjust its value across [0,1] with increments of 0.1 on the Facebook dataset. 
When $\gamma$ is set to 0, it indicates that the rationale loss term in RGMN is inactive.
In constast, when $\gamma$ is set to 1, the model optimizes RGMN based solely on the rationale loss term. 
As shown in Figure \ref{gamma}, when $\gamma$ is set to 0, both AUROC and AUPRC perform poorly, indicating that 
our rationale-aware Gaussian mixture normalizing flow plays a crucial role in RagGAD. 
However, when $\gamma$ is set to 1, despite the absence of guidance from the non-rationale loss term, 
the model still performs relatively well, achieving 93.82 and 29.08 for AUROC and AUPRC, respectively. 
When all loss terms are enabled and $\gamma$ is set to 0.7 to balance rationale and non-rationale losses, 
RagGAD performs optimally, achieving AUROC and AUPRC of 98.62 and 39.33, respectively.  

\begin{table*}[t]
\vspace{-0.6cm}
\centering
\setlength{\tabcolsep}{0.08cm} 
\renewcommand{\arraystretch}{0.7} 
\caption{The effect of parameters $\lambda_1$, $\lambda_2$, and $\lambda_3$ on different datasets.}
\vspace{-0.3cm}
\resizebox{0.85\textwidth}{!}{
\begin{tabular}{cc|cc|cc|cc|cc|cc|cc|cc}
\toprule
\rowcolor[HTML]{F0F0F0} & &  \multicolumn{2}{c|}{\textbf{BlogCatalog}}  &  \multicolumn{2}{c|}{\textbf{Amazon}} &   \multicolumn{2}{c|}{\textbf{Facebook}} &   \multicolumn{2}{c|}{\textbf{Reddit}}  &   \multicolumn{2}{c|}{\textbf{YelpChi-all}}  &   \multicolumn{2}{c|}{\textbf{T-Finance}}  &  \multicolumn{2}{c}{\textbf{Avg}}  \\  \cline{3-16}
\rowcolor[HTML]{F0F0F0}   \multicolumn{2}{c|}{\multirow{-2}{*}{\textbf{Dataset}}} &  \textbf{AUROC} & \textbf{AUPRC}   &  \textbf{AUROC} & \textbf{AUPRC}   & \textbf{AUROC} & \textbf{AUPRC}      & \textbf{AUROC} & \textbf{AUPRC}  & \textbf{AUROC} & \textbf{AUPRC} & \textbf{AUROC} & \textbf{AUPRC}  & \textbf{AUROC} & \textbf{AUPRC} \\ \midrule

& 0.001 & 77.10 & 35.11 & 88.57 & 67.94 & 89.63 & 25.60 & 63.33 & 5.60 & 62.40 & 22.37 & 91.76 & 59.27 & 78.30 & 36.65 \\
& 0.01  & 79.50 & 34.95 & 90.84 & 70.80 & 91.90 & 37.55 & \cellcolor{color1}65.62 &  \cellcolor{color1}5.79 & 62.80 & 22.79 & 89.73 & 60.37 & 79.56 & 39.69 \\
$\lambda_1$ & 0.1 & 86.31 & 45.47 & 91.85 & 74.85 & \cellcolor{color1}98.62 & \cellcolor{color1}40.26 & 63.49 & 5.35 & 62.86 & 22.79 &  \cellcolor{color1}92.98 & 68.37 &  \cellcolor{color1}81.35 &  \cellcolor{color1}39.91 \\
& 1     & \cellcolor{color1}87.32 & 46.32 & \cellcolor{color1}92.04 & \cellcolor{color1}77.67 & 94.10 & 30.58 & 63.60 & 5.59 & \cellcolor{color1}63.12 & 23.33 & 91.20 &  \cellcolor{color1}74.81 & 80.88 & 39.17 \\
& 10    & 77.04 & \cellcolor{color1}  36.41 & 91.93 & 75.32 & 90.20 & 22.18 & 60.75 & 5.15 & 62.77 &  \cellcolor{color1}23.36 & 84.77 & 47.46 & 77.12 & 35.19 \\

\midrule

& 0.001 & 86.74 &  \cellcolor{color1}47.60 &  \cellcolor{color1}93.07 & 76.20 & 93.88 & 26.70 & 64.28 & 6.31 & 63.41 & 22.63 & 89.28 & 55.26 & 80.96 & 40.06 \\
& 0.01  & 86.88 & 46.68 & 91.73 &  \cellcolor{color1}78.64 & 98.60 & 34.74 & 64.53 & 6.49 &  \cellcolor{color1}64.32 &  \cellcolor{color1}23.55 & 89.41 & 56.00 & 81.66 &  \cellcolor{color1}41.96 \\
$\lambda_2$ & 0.1 &  \cellcolor{color1}87.32 & 46.20 & 91.92 & 74.38 & 99.03 &  \cellcolor{color1}37.36 &  \cellcolor{color1}65.99 &  \cellcolor{color1}6.73 & 62.94 & 22.50 & 91.02 & 66.99 &  \cellcolor{color1}82.09 & 41.34 \\
& 1     & 86.96 & 46.70 & 91.15 & 74.00 &  \cellcolor{color1}99.08 & 31.41 & 63.03 & 5.58 & 62.15 & 22.30 & 91.50 &  \cellcolor{color1}86.43 & 81.24 & 40.69 \\
& 10    & 87.13 & 46.00 & 92.91 & 77.01 & 96.40 & 31.87 & 63.04 & 5.11 & 62.94 & 22.45 &  \cellcolor{color1}93.45 & 67.30 & 81.51 & 40.65 \\

\midrule

& 0.001 & 84.55 & 47.05 & 92.12 & 73.02 & 89.50 & 20.23 & 61.27 & 5.15 & 63.19 &  \cellcolor{color1}23.06 & 78.40 & 38.55 & 77.80 & 36.92 \\
& 0.01  &  \cellcolor{color1}87.01 & 47.40 & 91.45 & 76.40 & 94.02 & 24.84 &  \cellcolor{color1}65.26 & 5.60 &  \cellcolor{color1}64.38 & 22.70 & 82.28 & 53.78 & 80.26 & 38.93 \\
$\lambda_3$ &  0.1 & 86.32 & 47.10 & 91.62 &  \cellcolor{color1}78.68 & 95.90 & 26.00 & 65.00 &  \cellcolor{color1}6.00 & 63.04 & 22.18 &  \cellcolor{color1}93.50 &  \cellcolor{color1}77.00 &  \cellcolor{color1}81.90 &  \cellcolor{color1}40.12 \\
& 1     & 86.80 & 46.30 & 91.30 & 73.60 &  \cellcolor{color1}99.20 &  \cellcolor{color1}36.40 & 63.50 & 5.80 & 62.70 & 22.95 & 88.10 & 55.20 & 80.93 & 39.71 \\
& 10    & 85.60 &  \cellcolor{color1}47.90 &  \cellcolor{color1}92.80 & 74.60 & 96.10 & 26.90 & 62.00 & 5.70 & 62.90 & 22.50 & 82.50 & 47.10 & 80.00 & 38.45 \\

\bottomrule
\end{tabular}}
\label{lambdas}
\end{table*}

\begin{figure}[t]
\vspace{-0.4cm}
\centering
\includegraphics[width=0.49\textwidth]{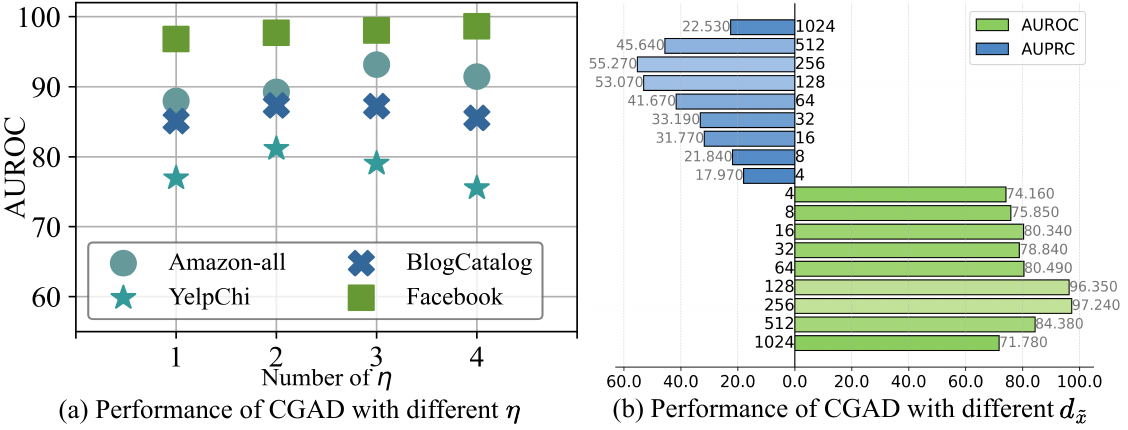} 
\vspace{-0.75cm}
\caption{Results of parameters sensitivity analysis. (a) shows the effects of different $\eta$, 
(b) shows the influence of embedding dimension $d_{\tilde{x}}$ on the ACM dataset.  
}   
\label{eta_and_dx}
\vspace{-0.4cm}
\end{figure}
 
\hspace{1em}{\emph{d) Sensitivity Analysis of $\eta$.}    
The hyperparameter $\eta$ defines the number of coupling layers in NCNF.  
We evaluate performance of RagGAD under varying $\eta$, with the results presented in Figure \ref{eta_and_dx}(a).
Overall, RagGAD demonstrates robustness to changes in $\eta$ across both small-scale graphs (BlogCatalog, Facebook, and YelpChi) 
and the large-scale graph (Amazon-all).
However, 
the highest AUROC is achieved at $\eta=1$ for small-scale graphs, 
while for large-scale graphs, the optimal setting is $\eta=2$. 
To ensure stable performance across different graph sizes while minimizing parameter redundancy, 
we set $\eta=2$ as the default value for all datasets.

\hspace{1em}{\emph{e) Sensitivity Analysis of $d_{\tilde{x}}$.} 
The hyperparameter $d_{\tilde{x}}$ in Eq.(\ref{projection}) denotes the projected attribute dimension for each node. 
Following GADAM \cite{chen2024boosting}, for datasets with low-dimension attributes, we set $d_{\tilde{x}}$ to 64. 
For high-dimension attribute datasets, to investigate the impact of $d_{\tilde{x}}$ on the performance of RagGAD, 
we vary $d_{\tilde{x}}$ within the range of 4 to 1024 on the ACM dataset, increasing it in powers of 2.
As shown in Figure \ref{eta_and_dx}(b), the performance improves as $d_{\tilde{x}}$ increases but starts to decline when it exceeds 256.
To strike a balance between effectiveness and computation efficiency,  
we set $d_{\tilde{x}}$  to 128 for all high-dimensional attribute datasets, including BlogCatalog, ACM, and Facebook.

\hspace{1em}{\emph{f)  Sensitivity Analysis of $\lambda_1$, $\lambda_2$, and $\lambda_3$.}
We perform a grid search to tune these hyperparameters separately for each dataset, with the search range set between 0.001 and 1. 
As shown in Table \ref{lambdas}, for $\lambda_1$, the optimal performance is obtained in the range of 0.01 to 1. 
$\gamma_1$ controls the strength of loss term $\mathcal{L}_{rgmn}$, which is the fundamental loss of RagGAD, 
the best performance is obtained when $\gamma_1=0.1$.
$\gamma_2$ and $\lambda_3$ control the contributions of the attribute reconstruction loss $\mathcal{L}_{rcon}$
and the rationale sparsity loss $\mathcal{L}_{spr}$, respectively. 
From Table \ref{lambdas}, we observe that $\lambda_2$ and $\lambda_3$ exhibit higher sensitivity compared to $\lambda_1$.
Specifically, $\lambda_2$ remains relatively stable in the range of 0.001 to 0.1, with a notable decrease in accuracy
observed when it exceeds 0.1. 
$\lambda_3$ remains relatively stable between 0.01 and 1, while the performance of RagGAD is suboptimal when $\lambda_3$ falls below 0.01.
This is mainly because setting $\lambda_2$ too high may cause the model 
to prioritize attribute reconstruction at the expense of accurately disentangling rationales. 
Conversely, setting $\lambda_3$ too low leads RagGAD to ignore the sparsity of rationales, 
resulting in the misclassification of weak non-rationale correlations as rationales.

\begin{table}[t] 
\vspace{-0.4cm}
\centering
\caption{The model complexity, parameter count, and training and testing time (in seconds) of different baselines. 
$R$ denotes the number of evaluation rounds,
$c$ denotes the number of nodes in the local subgraph,
$k$ is the number of global affinity truncation graph neighbors \cite{zhang2025gctam},
and $\omega$ is the average node degree of $\mathcal{G}$ \cite{chen2024boosting}. }
\vspace{-0.3cm}
\resizebox{0.48\textwidth}{!}{
\begin{tabular}{r|c|c|c|c}
\toprule
\rowcolor[HTML]{F0F0F0}  \multicolumn{1}{c|}{\textbf{Method}} & \textbf{Overall Complexity} & \textbf{Parameters} & \textbf{Train} & \textbf{Test}  \\
\midrule
DOMINANT\cite{ding2019deep}   & $\mathcal{O}(\mathcal{E} + N^2)$                                 & 15769  & 67.549 &  0.283 \\
CoLA\cite{liu2021anomaly}     & $\mathcal{O}(cNR(c + \omega))$                                   & 5762   & 285    &  2.740  \\
GADAM\cite{chen2024boosting}  & $\mathcal{O}(\mathcal{E} + N)$                                   & 4226   & 6.418  &  0.029   \\
TAM\cite{qiao2024truncated}   & $\mathcal{O}(\mathcal{E}d_x + \mathcal{E}d_h + N^2)$             & 105091 & 1039   &  217 \\
CoCo \cite{wang2025context}   & $\mathcal{O}(\mathcal{E} + Nd_{\tilde{x}}^2 + N^2d_{\tilde{x}})$ & 12283  & 11.26  &  0.034 \\   
HUGE\cite{pan2025label}       & $\mathcal{O}(N^2d_h)$                                            & 36355  & 18.38  &  13.45 \\
GCTAM \cite{zhang2025gctam}   & $\mathcal{O}(Nd+ \mathcal{E}d + Nk)$                             & 72320  & 41.72  &  0.3876\\
\multicolumn{1}{c|}{RagGAD}   & $\mathcal{O}(N^2d_{\tilde{x}})$                                  & 21439  & 87.72  &  0.6488\\
\bottomrule
\end{tabular}}
\label{Complexity}
\vspace{-0.35cm} 
\end{table}

\subsection{Efficiency and Complexity Analysis}
We evaluate efficiency from model complexity, parameter count, and empirical runtime, 
as summarized in Table \ref{Complexity}. 
All runtimes are measured on the Amazon dataset, 
where the training time corresponds to 100 epochs and the testing time reflects standard inference cost.

In terms of complexity, most baselines scale linearly with $\mathcal{E}$ or $N$, such as GADAM and GCTAM, 
while methods modeling pairwise node interactions introduce quadratic terms $\mathcal{O}(N^2)$, 
including DOMINANT, TAM, and HUGE. 
RagGAD falls into this category with complexity $\mathcal{O}(N^2 d_{\tilde{x}})$, 
due to global node-wise interaction and rationale disentanglement in the ARD module. 
Compared with local or sampling-based methods, this design captures long-range dependencies 
and high-order structural correlations, which are critical for accurate anomaly characterization.
Despite the quadratic complexity, RagGAD maintains a moderate parameter size of 21,439, 
which is significantly smaller than TAM and comparable to GCTAM and HUGE. 
This indicates that the additional cost stems from dense computation rather than over-parameterization, 
reflecting efficient parameter utilization.
Empirically, RagGAD achieves competitive efficiency. 
Its training time of 87.72 seconds is much lower than TAM and CoLA, 
and its inference time of 0.6488 seconds remains practical, outperforming several high-complexity methods. 
This suggests that the theoretical complexity does not directly translate into prohibitive runtime, 
as optimized matrix operations and parallelism mitigate the overhead in practice.

Overall, RagGAD introduces additional computation for global dependency modeling and rationale disentanglement, 
leading to improved representation robustness, 
while maintaining a favorable balance between effectiveness and efficiency without excessive parameter or runtime overhead.

\section{Conclusion}
In this paper, we propose RagGAD, a novel unsupervised graph anomaly detection framework that disentangles
robust rationales from non-rationale correlations in node interrelationships. 
By further decomposing rationales into robust and fragile components, RagGAD captures the underlying mechanisms governing normal 
behavior while identifying anomalies through deviations in non-rationale interactions. 
To model the complex distributions of both normal and abnormal nodes, 
we incorporate a node-level rationale-aware conditional Gaussian mixture normalizing flow that enhances sensitivity 
to the diversity of normal patterns and mitigates the homophily trap. Extensive experiments on benchmark datasets demonstrate
that RagGAD outperforms state-of-the-art baselines, confirming its effectiveness in addressing the challenges of unsupervised 
graph anomaly detection. Future work will focus on optimizing the model to improve efficiency without sacrificing performance, 
and expanding its applicability to dynamic and heterogeneous graph scenarios.


\bibliographystyle{IEEEtran}
\bibliography{example_paper}

\end{document}